\documentclass[11pt]{article}

\usepackage[utf8]{inputenc} 
\usepackage[T1]{fontenc}    
\usepackage{hyperref}       
\usepackage{url}            
\usepackage{amsfonts}       
\usepackage{nicefrac}       
\usepackage{microtype}      
\usepackage{xcolor}         

\usepackage{algorithm,algpseudocode}
\usepackage{float}

\usepackage{enumitem}
\usepackage{algpseudocode}
\usepackage{amsmath,amsfonts,amssymb}     
\usepackage{graphicx}
\usepackage{mathabx}
\usepackage{csquotes}
\usepackage{bm}
\usepackage{subcaption}
\usepackage{comment}
\usepackage{xcolor}
\usepackage{tabu}
\usepackage{bbm}
\usepackage{wrapfig}
\usepackage{steinmetz}
\usepackage{cancel}

\graphicspath{{./figure/}}

\usepackage{tikz}
\usetikzlibrary{shapes.geometric} 
\usetikzlibrary{arrows, backgrounds, calc, hobby, positioning}

\tikzset{vertex style/.style={
    draw=#1,
    thick,
    fill=#1!70,
    text=white,
    ellipse,
    minimum width=1cm,
    minimum height=0.75cm,
    font=\footnotesize,
    outer sep=2pt,
  },
  text style/.style={
    sloped,
    text=black,
    font=\footnotesize,
    above
  }
}

\tikzset{vertex style1/.style={
    draw=#1,
    thick,
    fill=#1!70,
    text=white,
    circle,
    minimum width=.5cm,
    minimum height=0.5cm,
    font=\footnotesize,
    outer sep=1pt,
  },
  text style/.style={
    text=black,
    font=\footnotesize,
    above
  }
}

\newtheorem{theorem}{Theorem}
\newtheorem{lemma}{Lemma}
\newtheorem{ex}{Example}
\newtheorem{defn}{Definition}
\newtheorem{as}{Assumption}

\newtheorem{prop}{Proposition}
\newtheorem{cor}{Corollary}
\newtheorem{remark}{Remark}

\newcommand{\mbZ}{\mathbb{Z}}
\newcommand{\I}{\mathcal{I}}
\newcommand{\mbN}{\mathbb{N}}
\newcommand{\mbR}{\mathbb{R}}
\newcommand{\mbC}{\mathbb{C}}

\newcommand{\independent}{\perp \!\!\! \perp}

\def\X{{\mathbf X}}
\def\x{{\mathbf x}}

\def\A{{\mathbf A}}
\def\B{{\mathbf B}}

\def\E{{\mathbf E}}

\def\H{{\mathbf H}}
\def\I{{\mathbf I}}

\def\N{{\mathbf N}}

\def\Y{{\mathbf Y}}
\def\Z{{\mathbf Z}}

\def\W{{\mathbf W}}

\def\z{{\mathbf z}}

\def\1{{\mathbbm 1}}

\def\mP{{\mathbb P}}

\def\mB{{\mathcal B}}
\def\mC{{\mathcal C}}
\def\mE{\Vec{{\mathcal E}}}
\def\uE{{\mathcal E}}
\def\mF{{\mathcal F}}

\def\mG{{\mathcal G}}

\def\mK{{\mathcal K}}

\def\mM{{\mathcal M}}
\def\mR{{\mathcal R}}
\def\mS{{\mathcal S}}
\def\mSP{\mS{\mathcal P}}

\def\mX{{\mathcal X}}
\def\mX{{\mathcal X}}

\def\pspace{(\Theta,\mF,\mP)}

\def\hX{ {\widehat \X}}

\def\oi{ {\overline{i}}}

\newcommand\hatt[1]{\widehat{#1}}

\def\<{{\langle}}
\def\>{{\rangle}}

\newcommand\dsep[2]{\mathrm{d\text{-}sep}_{#1}(#2)}

\title{Causal Structure Recovery of Linear Dynamical Systems: An FFT based Approach}

\author{Mishfad Shaikh Veedu\footnote {Department of Electrical Engineering, University of Minnesota Twin Cities}, James Melbourne
\footnote { Department of Probability and Statistics, Centro de Investigación en Matemáticas (CIMAT)}, and Murti V. Salapaka$^*$}

\begin{document}

\maketitle

\begin{abstract}
Learning causal effects from data is a fundamental and well-studied problem across science, especially when the cause-effect relationship is static in nature. However, causal effect is less explored when there are dynamical dependencies, i.e., when dependencies exist between entities across time. Identifying dynamic causal effects from time-series observations is computationally expensive when compared to the static scenario \cite{krishnan2023simplicial,pavlovic1999dynamic}. We demonstrate that the computational complexity, of recovering the causation structure, for the vector auto-regressive (VAR) model (see \cite{krishnan2023simplicial,peters2013causal}) is $O(Tn^3N^2)$, where $n$ is the number of nodes, $T$ is the number of samples, and $N$ is the largest time-lag in the dependency between entities. We report a method, with a reduced complexity of $O(Tn^3 \log N)$, to recover the causation structure that  uses fast Fourier transform (FFT) to obtain frequency-domain representations of time-series. Since FFT accumulates all the time dependencies on every frequency, causal inference can be performed efficiently by considering the state variables as random variables at any given frequency. We additionally show that, for systems with interactions that are linear and time-invariant, do-calculus machinery can be realized in the frequency-domain resulting in versions of the classical single-door (with cycles), front-door and back-door criteria. Moreover, we demonstrate, for a large class of problems, graph reconstruction using multivariate Wiener projections results in a significant computational advantage with  $O(n)$ complexity over reconstruction algorithms such as the PC algorithm which has $O(n^q)$ complexity, where $q$ is the maximum neighborhood size. This advantage accrues due to some remarkable properties of the phase response of the frequency dependent Wiener coefficients which is not present in any time-domain approach. Concentration bounds are utilized to obtain non-asymptotic estimates of how well the Wiener coefficients can be determined, delineating the effect of data-size and accuracy of the estimates. Using the machinery developed, we also define and analyze two conditional independence notions for dynamically dependent stochastic processes and showcase the effectiveness of the framework developed by obtaining a simple proof of the single-door criteria in the presence of directed cycles. Finally, we verify the results with the help of simulations.
\end{abstract}

\section{Introduction and literature survey}
Causal identification from data is an active and important research area relevant to multiple domains including climate science \cite{perez2018causal}, economics \cite{carfi2008financial}, neuroscience \cite{ramirez2021coupling}, and biology \cite{Causality_genome}. There is a considerable prior art on causation especially when the interactions are static in nature (see \cite{pearl2016causal,peters2017elements,spirtes2000causation} and the references therein). In the case when entities can be modeled as random variables with an underlying joint probability distribution, without the possibility of actively intervening in the system, it is not possible in general to reconstruct the causal graph from observational data alone. Indeed, Markov equivalent graphs that capture the same set of conditional independence (CI) relations can be determined from data \cite{spirtes2000causation}. Some recent works on recovering a unique causal graph when the underlying data generative system is assumed to have more structure is presented in \cite{peters2013causal,shimizu2006linear}. Considerable prior art and algorithms exist for recovering the causal graph of entities modeled as random variables, when interventions on the system are possible \cite{spirtes2000causation}.  In a related problem of estimating the causal effect given the causal graph, the goal is to estimate the effect of a variable on another.  Here, a powerful framework of do-calculus can be employed to ascertain the set of variables that need to be used to determine the total and direct effect of a variable on another which result in criteria such as the single-door, back-door and front-door criteria \cite{maathuis2009estimating,pearl2016causal}

Causal inference is more challenging in the presence of dynamical (across time)  dependencies \cite{costanzo2020data,krishnan2023simplicial,peters2013causal}. 
One of the established techniques for identifying the causation is Granger causality \cite{granger1969investigating}, which utilizes how well a set of variables can be employed to estimate a given variable while utilizing the time-arrow.  Granger causality based approaches necessitate presence of delays in the dynamic dependencies. Such an assumption is rendered problematic when the data is collected at slower rates than the time-constants at which the dynamics evolve, thus precluding a number of practical scenarios.  Another approach in handling dependencies across times is to consider the series at every time instant as a random variable thereby mapping the problem to a static version; the difficulties with such an approach stem from the lack of information on the size of the horizon in the past and future to be considered and the combinatorial explosion of the number of variables that entail \cite{ghahramani2006learning_dynamic,lohmann2012critical}. In a class of approaches for unveiling dynamic dependencies,  models that include vector auto regressive (VAR) models \cite{krishnan2023simplicial,peters2013causal}, additive noise models (ANM) \cite{costanzo2020data,peters2013causal} and neural network based models (NNM) \cite{moraffah2021causal},  of how the data is being generated is assumed and the causal graph structure is recovered via an estimation of model parameters. In another recent line of work, the main gist of the approach is to exploit the asymmetry in the relationships of a dependency and the inverse of the dependency.  These asymmetries can result from nonlinear maps or from linear filters with inverses that have discernable differences that, for example, are characterizable via power spectral densities \cite{besserve2022cause,shajarisales2015telling}.  Here, it is to be noted that works such as \cite{besserve2022cause,shajarisales2015telling} try to infer cause and effect relationship between two variables; however, the causal graph structure recovery entails determining how the influence flows from a variable to another with the possibility of intermediate variables between the cause and the effected variables.  We emphasize that effective methodologies for causal inference for systems with dynamic interactions remain much less developed than their static dependency counterparts. 

  Several recent works have bridged  the  causality literature (\cite{spirtes2000causation})  with multivariate projections of dynamically related time-series data
\cite{costanzo2020data,materassi2016graphoid,Salapaka_signal_selection}. \cite{Salapaka_signal_selection} extended the single-door criterion \cite{pearl2016causal} to include cycles and proposed a similar criterion called revolving-door criterion.  It was shown in \cite{materassi_tac12} that the Wiener projections can be employed to recover the moral graph of entities interacting via  linear time-invariant dynamics.

The main contributions of the article are summarized below. This article develops a framework for system with dynamical interactions using a frequency domain approach. Here, the main steps taken for reconstruction of the moral graph (which recovers the kin structure) is: transform time-series of each entity (modeled as a stochastic process) to  frequency domain using Fast Fourier Transform (FFTs), perform multivariate  projections in the {\it frequency domain}, and construct an undirected graph using the non-zero entries of the obtained multivariate projection filter. Under the assumption that the data generative model is linear and time-invariant it is shown that the above process recovers the moral graph. With the standard assumption of faithfulness \cite{spirtes2000causation}, a process akin to the Peter-Clark (PC) algorithm is employed to recover the essential graph with better computational efficiency, which exploits the phase properties of the complex valued estimates. When interventions are possible, it is shown that the causal graph can be recovered.  It is demonstrated that the computational complexity for recovering the causal structure   using the projections in frequency domain is $O(Tn^3 \log N)$ compared to the conventional projections using  time-domain which are shown to have  a complexity of   $O(Tn^3N^2)$; $n$, $T$ and $N$ are the number of nodes, number of samples, and the largest time-lag in the dependency between entities, respectively.  Two notions of independence for stochastic processes are developed  with  guarantees on the convergence of the FFT model to the continuous frequency model provided. These notions help generalize and streamline the framework based on joint probability distributions in the static case to the dynamic case with the mechanics of do-calculus made possible in the dynamics case. For estimating Wiener coefficients from time-series data, a non-asymptotic concentration bound and a sample complexity analysis are also  provided. Furthermore,  for causal-effect estimation for LTI dependencies, back-door, single-door (with cycles) and front door criteria are obtained with proofs that are rendered straightforward in the frequency domain.

\emph{Notations:} Capital letters, $X$, denote either random variables or sets, the usage will be clear from the context; bold letters denote vectors or matrices. For any time signal, $x:=\{x(t):t\in \mbZ\}$, $\hatt{x}$ denotes the Fourier transform of $x$, $\hatt{x}(e^{j\omega}):=\sum_{n=-\infty}^\infty x(n) e^{-j\omega n}$; for stochastic processes, $\X$, $\Phi_\X(e^{j\omega})$ denotes the power spectral density (PSD) of $\X$; $\mbR,\mbC,\mbZ$ denote the set of real numbers, complex numbers and integers respectively; $[N]:=\{0,\dots,N-1\}$; for a complex number, $x=x_R+ix_I$, $\angle x$ denotes $\arctan(x_I/x_R)$; $\Omega$ denotes the interval $[0,2\pi]$ and $\Omega_N$ denotes $\{0,2\pi/N,\dots,2\pi(N-1)/N\}$; for a transfer function, $\hatt{x}$, $\hatt{x} \neq 0$ means $\hatt{x}$ is not identically zero; $\|\hatt{x}\|_{\ell_\infty(\Omega_N)}$ denotes $\sup_{\omega \in \Omega_N}\hatt{x}(\omega)$; in a given directed graph,  $Pa(i), Ch(i), Sp(i)$ denotes the set of parents, children, and spouses of node $i$ respectively. $\ell^1$ denotes the space of absolutely summable sequences.

\section{Preliminaries}
\subsection{Linear Dynamic Influence Models (LDIM)}
Consider a network with $n$ nodes, each node  $i\in\{1,\dots,n\}$ having the time-series measurements, ${x}_i$, governed by the 
{linear dynamical influence model (LDIM)},
\begin{align}
\label{eq:LDIM} 
	\hatt{\X}(e^{j\omega}) &= \hatt{\H}(e^{j\omega})\hatt{\X}(e^{j\omega})+\hatt{\E}(e^{j\omega}), ~\forall \omega \in [0,2\pi],
\end{align}
where $\hatt{\X}=[\hatt{X}_1,\dots,\hatt{X}_n]^T$, and $\hatt{\E}=[\hatt{E}_1,\dots,\hatt{E}_n]^T$ are the exogenous noise source with $E_i,E_j$ jointly wide sense stationary for $i\neq j$. $\hatt{\H}$ is a well posed transfer function, i.e., every submatrix of $(\I-\hatt{\H}(e^{j\omega}))$ is invertible as well as every entry of $(\I-\hatt{\H}(e^{j\omega}))^{-1}$ is analytic, and $\hatt{\H}_{ii}(e^{j\omega})=0$ for every $\omega\in \Omega$.  $\Phi_\E(e^{j\omega})$ is positive definite and diagonal for every $\omega$. For notational convenience, $\hatt{x}(e^{j\omega})$ is replaced with $\hatt{x}(\omega)$ henceforth. Notice that the LDIM \eqref{eq:LDIM} can be represented in time-domain using the following linear time invariant model,
\begin{align}
\label{eq:convolution_model}
{{\X}}(k) &= \sum_{l=-\infty}^{\infty} {\mathbf{H}}(l){{\X}}(k-l)+{{\E}}(k),
\end{align}
    where ${{\E}}(k) = [{E}_1(k),\dots,{E}_n(k)]^T$, and ${\X}(k) = [{X}_1(k),\dots,{X}_n(k)]^T$. The LDIM \eqref{eq:LDIM} can be represented using a graph $G=(V,\mE)$, where $V=\{1,\dots,n\}$ and $\mE=\{(u,v):H_{vu}\neq 0\}$, that is, there exists an edge $u\rightarrow v$ in $\mE$ if $\hatt{H}_{vu}(\omega) \neq 0$ for some $\omega \in [0,2\pi]$. $u$ is said to be a kin of $v$ in $G$, denoted, $u\in kin_G(v)$, if at least one of the following exist in $\mE$: $u\rightarrow v$, $v\rightarrow u$, $u\rightarrow i \xleftarrow{}v$ for some $i \in V$. $kin_G(j)$ denotes the set of kins of $j$ in $G$.
\subsection{Frequency Discretization}
The  LDIM in \eqref{eq:LDIM} can be  sampled at frequencies $\omega$ in $\Omega_N=\left\{\omega_0,\dots,\omega_{N-1}\right\}$, where $\omega_k:=\frac{2\pi k}{N}$, thus converting the continuous frequency model to practical model involving finite set of frequencies. In time-domain, the sampled set of relations of  \eqref{eq:LDIM} is equivalent to  the finite impulse response model described by,
\begin{align}
\label{eq:circular_convolution_model}
{{\X}}(n) &= \sum_{l=0}^{N-1} 
{\mathbf{H}}(l) {\X}(n-l)_{mod ~N}+{\E}(n), ~ n=0,\dots,N-1,
\end{align}
where $\mathbf{X}(n)$ and $\E(n)$ are periodic with period $N$ and $\H(l)$ is non-zero for at most $N-1$ lags.
Let $  \hatt{\mathbf{X}}(e^\frac{j2\pi k}{N})=\frac{1}{\sqrt{N}} \sum_{n=0}^{N-1} \mathbf{X}(n) e^{-2\pi kn/N}$, $k=0,\dots,N-1$ which provides the Discrete Fourier Transform (DFT) of the sequence $X(n).$
Similar expressions hold for $\H(k)$ and $\E(k)$ and their corresponding DFTs. The directed graph associated with  \eqref{eq:circular_convolution_model} is, $\breve{G}=(V,\breve{E})$ is defined such that the edge $u\rightarrow v$ exists in $\breve{E}$ if $\hatt{H}_{vu}(e^\frac{j2\pi k}{N}) \neq 0$ for some $k$. A pertinent question here is: when and under what conditions \eqref{eq:circular_convolution_model} converges to \eqref{eq:convolution_model}. It is well known that the Fourier coefficients can be approximated via discretization. The following result, with proof included for completeness in the supplementary material, gives a uniform and explicit convergence rate for signals with bounded variation.
\begin{theorem}
\label{thm:convergence}
    For a function $\hatt{f}$ of bounded variation $V$, on $[0,2\pi]$ and $N\geq 1$, the estimation of $f(n)=\int_{0}^{2\pi} \hatt{f}(\omega) e^{j \omega n} d\omega$ given by $f^{(N)}(n)=\frac{1}{N}\sum_{k=0}^{N-1} e^{j2\pi nk/N} f(2 \pi k/N)$ for $|n| \leq \sqrt{N}$ and zero otherwise, satisfies $\|f-\hatt{f}^{(N)}\|_{\ell_\infty(\Omega)} \leq C/\sqrt{N}.$
\end{theorem}
\textbf{Proof:}
    See Appendix \ref{app:convergence}. \hfill
\subsection{Graph definitions}


Consider a directed graph $G=(V,{
\mE})$, where $V=\{1,\dots,n\}$. A {\it chain} from node $i$ to node $j$ is an ordered sequence of edges in ${\mE}$, $((\ell_0,\ell_1),(\ell_1,\ell_2)\dots,(\ell_{n-1},\ell_n))$, where $\ell_0=i$, $\ell_n=j$, and $(\ell_i,\ell_{i+1}) \in \mE$. Topology, $top(G)$, of a directed graph, $G=(V,\mE)$, is an undirected graph, $top(G)=(V,\uE)$ where an edge $(u,v) \in \uE$ if either $(u,v)\in \mE$ or $(v,u)\in \mE.$
A {\it path} from node $i$ to node $j$ in the directed graph $G$ is an ordered set of  edges $((\ell_0,\ell_1),(\ell_1,\ell_2)\dots,(\ell_{n-1},\ell_n))$ with $\ell_0=i$, $\ell_n=j$ in its topology, $(V,\uE)$, where  $\{\ell_i,\ell_{i+1}\} \in \uE$.

A path of the form $((\ell_0,\ell_1),\dots,$ $(\ell_{n-1},\ell_n))$ has a collider at $\ell_k$ if $\ell_k \in Ch(\ell_{k-1}) \cap Ch(\ell_{k+1})$. That is, there exists directed edges $\ell_{k-1} \rightarrow \ell_{k}\xleftarrow[]{} \ell_{k+1}$ in $G$. Consider disjoint sets $X,Y,Z \subset V$ in a directed graph $G=(V,\mE)$. Then, $X$ and $Y$ are \textbf{\emph{d-separated}} given $Z$ in $G$, denoted $\dsep{G}{X,Z,Y}$ if and only if every path between $x \in X$ and $y\in Y$ satisfies at least one of the following: 1) The path contains a non-collider node $z\in Z$. 2) The path contains a collider node $w$ such that neither $w$ nor the descendants of $w$ are present in $Z$. In-nodes of $i$ is defined as $\{j:(i,j)\in \mE\}$.

\subsection{Graph/Topology learning}
 We first present a method for estimating the $i^{th}$ time-series from another set of time-series. It is well known that the optimal estimate, $\widetilde{X}_i$, (in the minimum mean square sense) of the process $X_i$ from the rest of the time-series in $\overline{i}=V\setminus \{i\},$  satisfies $\hatt{\widetilde{X}}_i(\omega)=W_{i\cdot \overline{i}}(\omega)\hatt{X}_{\overline{i}}(\omega)$ where  $W_{i\cdot \overline{i}}(\omega)=\Phi_{\overline{i},\overline{i}}^{-1}(\omega)\Phi_{i,\overline{i}}(\omega)$, with  $\Phi_{i,j}(\omega):=\sum_{k=-\infty}^\infty R_{ij}(k)e^{-j\omega k}$, $R_{ij}(k):=\mathbb{E}[X_i(k)X^T_i(0)]$. Here $W_{i\cdot C}(\omega)$ is multivariate Wiener filter obtained by projecting $X_i$ onto the set of processes in $C$ \cite{kailath2000linear}.  The entry of $W_{i\cdot C}$ corresponding to $j^{th}$ node in the set $C$, is denoted by $W_{i\cdot C}[j]$. For notational convenience, the Wiener coefficient (vector of size $|\oi|$) in the projection of $X_i$ to $X_{\oi}$ is denoted by $W_i.$


\begin{lemma}
\label{lem:Wiener_kin_graph}  Consider a well-posed LDIM given by \eqref{eq:LDIM}. Let $W_i:=W_{i\cdot \oi}$ be the Wiener projection of $\X_i$ to $\X_{\oi}$. Then, $W_{i}[j](\omega)\neq 0$ if and only if $i \in kin_G(j)$.
\end{lemma}
\textbf{Proof:}
    See Appendix \ref{app:indep_G_implies_ind_omega}
    \begin{as} 
\label{As:phase}
If a node $k$ has multiple incoming edges in $G$, then for every pair of in-nodes $i,j$ of $k$, $\angle \hatt{H}_{ki}=\angle \hatt{H}_{kj}$, where $i,j,k \in V$.
\end{as}
\begin{as}
\label{as:imaginary_non-zero}
    If $\hatt{H}_{ij}(\omega)\neq 0$ then $\Im\{\hatt{H}_{ij}(\omega)\}\neq 0$, for every $i,j \in V$ and $\omega \in \Omega$.
\end{as}
\begin{lemma}
\label{lem:Wiener_imaginary} \cite{TALUKDAR_physics}
    Consider a well-posed LDIM given by \eqref{eq:LDIM} and satisfying Assumptions \ref{As:phase} and \ref{as:imaginary_non-zero}. Then for any $\omega \in \Omega$, $\Im\{W_i[j](\omega)\} \neq 0$ if and only if $(i,j) \in \mE$ or $(j,i) \in \mE$.
\end{lemma}
 We first briefly outline the  {\it time-domain based approach} in determining the multivariate Wiener filter coefficients. Define $m=|C|$, $\X_i:=[\X_i(T),\X_i(T-1),\dots,\X_i(N)]^T$  and $\mathbf{Y}_C:=[\X_C(T), \X_C(T-1),\dots,\X_C(N)]^T \in \mathbb{R}^{(T-N+1) \times |C|N}$, where $\X_C(k):=[\X_{c_1}(k),\X_{c_1}(k-1),\dots,\X_{c_1}(k-N+1),\X_{c_2}(k),\X_{c_2}(k-1),\dots,\X_{c_2}(k-N+1),\dots,\X_{c_m}(k),\X_{c_m}(k-1),\dots,\X_{c_m}(k-N+1) ] \in \mathbb{R}^{|C|N}$. Consider the least square formulation described by  $\beta_T:=\arg \min_{\beta\in \mbR^{N|C|} }\frac{1}{T}\|\X_i-\Y_C\beta \|_2^2$ obtains the Wiener filter coefficients $\beta_T$ for a finite time horizon of length $T.$  Let 
 \begin{align}
 \label{eq:Wiener_filter_time}
     \widetilde{\W}_{i\cdot C}:=\arg \min_{\beta\in \mbR^{N|C|} } \lim_{T\rightarrow \infty} \frac{1}{T}\|\X_i-\Y_C \beta\|_2^2.
 \end{align} Then, it can be shown that the Wiener filter $W_{i\cdot C}$ coincides with $\widetilde{\W}_{i\cdot C}.$ Here $\beta_T$  for sufficiently large $T$ is employed as a representative for the Wiener filter $W_{i\cdot C}.$ Non-asymptotic concentration bounds and sample complexity results for estimating Wiener coefficient $W_i$ from time-series data is provided in Section \ref{subsec:sample_complexity}.

We now outline a process of determining multivariate Wiener filter by first transforming every time-series to its frequency domain representation followed by {\it projections in the frequency domain.} Consider the $i^{th}$ time-series $X_i$ which is partitioned in $R$ segments with the $r^{th}$ segment denoted by $X_i^r$. Each segment consists of $N$ samples, for example,  $X_i^r=\{X_i^r(0),\ldots X_i^r(N-1)\}$.
Thus, the time series $X_i$ is given by, $\{(X_i^r(t))_{t=0}^{N-1}\}_{r=1}^R$. Using the $r^{th}$ segment of the  $X_i$ trajectory, given by $X_i^r(0),\dots,X_i^r(N-1)$, the FFT, $\hatt{\X}_i^r(e^\frac{j2\pi k}{N})=\frac{1}{\sqrt{N}} \sum_{n=0}^{N-1} \mathbf{X}_i^r(n) e^{-2\pi kn/N}$ is computed. Let $C=\{c_1,c_2,\ldots,c_m\} \subseteq V$ where $V$ is the set of time-series. Let $\hatt{\X}_C^r$ be the vector $\left[\begin{array}{cccc} 
\hatt{\X}_{c_1}^r & \hatt{\X}_{c_2}^r\ldots & \hatt{\X}_{c_m}^r\end{array}\right]^T$ which is the vector obatined by stacking the  Fourier coefficients obtained from the $r^{th}$ segments of the time-series in the set $C$. Let $\mX_C:=[\hatt{\X}_C^1,\dots,\hatt{\X}_C^R]^T$, where $C \subseteq V$.

It can be shown that (see \cite{doddi2022efficient})  
\begin{align}
\label{eq:Wiener_filter}
    W_{i\cdot C}(\omega):=\arg \min_{\beta \in \mbC^{|C|}}  \lim_{N,R\rightarrow \infty} \frac{1}{2R}\|\mX_i(\omega)-\mX_{C}(\omega)\beta\|_2^2, ~\omega \in [0,2\pi].
\end{align}
 Here $\beta_{RN}:=\arg \min_{\beta \in \mbC^{|C|}}   \frac{1}{2R}\|\mX_i(\omega)-\mX_{C}(\omega)\beta\|_2^2, ~\omega \in [0,2\pi]$ for sufficiently large $RN$ is employed to represent the Wiener filter $W_{i\cdot C}.$

 \subsection{Conditional Independence on a Set of Stochastic Processes and Factorization According to Directed Graphs}
A framework for  do-calculus for stochastic processes (SPs) requires a notion of independence in SPs and factorization according to conditional independence (CI). Towards this, we provide two notions of independence for SPs, the first one, motivated by the Kolmogorov extension theorem \cite{CShalizi_lec_notes}  is based on the finite dimensional distribution (FDD) in time-domain. The second notion is defined in the Fourier domain. Using each of the two CI notions, one can define factorization of SPs according to graphs, perform do-calculus and derive the single-door, front-door and back-door methods for direct and total effect identification.  The FDD based CI notions are developed in the   supplementary material due to space constraint. Here, in the main article, we focus on the independence in frequency-domain, which requires existence of Fourier transform.


\subsubsection{Conditional Independence in Stochastic Processes Interacting via LDIMs}
\label{subsec:CI_SP}


 An SP $X:=\{X(t):t \in I\}$, time indexed by $I=\mbZ$ (or $I=[N]$) on the probability space $\pspace$ is a map from $\Theta \times I \rightarrow \mbR^I$. Here, $\X_i$ is used to denoted the SP $\{X_i(t): t \in I\}$. 
For any $\theta \in \Theta$, $t\in I$, $X(\theta,t)$ denotes an instance of $X$ at time $t$. The set $X_\theta:=\{X(\theta,t)\}_{t\in I}$ is the \textbf{sample path} corresponding to the event $\theta\in \Theta$. We assume that $X_\theta \in \ell^1$. For every $\theta \in \Theta$, the discrete time Fourier transform of the sample path, $\widehat{X}_\theta: (\Theta,\mM) \mapsto \mbC^\Omega$ is defined as 
$\widehat{X}_\theta(\omega):=\sum_{t\in I}X(\theta,t) e^{-i\omega t}, ~\forall \omega \in \Omega$.
Then, $\hatt{X}(\omega):\Theta \rightarrow \mbC$ is a random variable and it is possible to define the independence in the frequency-domain as follows.

A set of stochastic processes $\{X_i\}_{i=1}^m$ on $\pspace$ are independent at $\omega \in \Omega$ if and only if 
\begin{align*}
    \mP\left( \bigcap_{i=1}^m \left\{ \widehat{X}_i(\omega) \in B_{i}\right\} \right)=\prod_{i=1}^m\mP\left(  \widehat{X}_i(\omega) \in B_{i} \right).
\end{align*}
\begin{defn}
[Conditional Independence]
\label{def:condi_inde_freq}
Consider the stochastic processes $X_1$, $X_2$, and $X_3$. Then $X_1$ is said to be conditionally independent of $X_2$ given $X_3$ at $\omega \in \Omega$ (denoted $X_1 \independent^{(\omega)} X_2 \mid X_3$) if and only if
\begin{align*}
    \mP\left( \bigcap_{i=1}^2 \left\{\widehat{X}_i(\omega) \in B_{i}\right\} \Bigm\vert \hatt{X}_3(\omega) \in B_{3} \right)=\prod_{i=1}^2\mP\left(  \widehat{X}_i(\omega) \in B_{i}\Bigm\vert \hatt{X}_3(\omega) \in B_{3} \right),
\end{align*}
where $B_i \in \mathcal{B}^2$; $\mathcal{B}^2$ is the Borel sigma algebra.
\end{defn}
In the LDIM \eqref{eq:LDIM}, $\hatt{\X}$, if it exists, is a linear transform of $\X$. Since FT-IFT pair is unique, the probability density functions (pdfs) are well-defined, if the pdf exists in time-domain. Then, $\hatt{\X}(\omega)$ can be considered as random vectors with an associated probability measure  $\mathbb{P}$ and a family of density functions $\{f^{(\omega)}\}_{\omega \in \Omega}$ on the set of nodes $V$. 

We now provide a definition of factorization at a specific frequency.
\begin{defn}[Factorization of SPs according to a directed graph at a specific frequency]
\label{def:factorization_SP_omega}
A set $V$ of stochastic processes is said to \textbf{factorize} according to a \textbf{directed} graph $G^{(\omega)}=(V,\mE^{(\omega)})$ at $\omega\in [0,2\pi]$ if and only if the probability distribution at $\omega$ can be factorized according to the graph $G^{(\omega)}$, i.e., 
{
\begin{align}
\label{eq:factorization_dist_graph_omega}
\mP\left(\hatt{X}_1(\omega),\dots,\hatt{X}_n(\omega)\right)=\prod_{i\in V} \mP\left(\hatt{X}_i(\omega) \mid \hatt{X}_{Pa_i}(\omega) \right),
\end{align}}
where $Pa_i$ denotes the parents of $i$ in $G^{(\omega)}$. If the pdf exists, then
\begin{align}
\label{eq:atomic_intervention_density_split}
f^{(\omega)}\left(\hatt{x}_1(\omega),\dots,\hatt{x}_n(\omega) \right)= \prod_{i=1}^n f^{(\omega)}\left(\hatt{x}_i(\omega) \mid \hatt{x}_{Pa_i}(\omega) \right).
\end{align}
Then, we can define a graph $G=(V,\mE)$, where $V=\{1,\dots,n\}$ and $\mE$ is such that the edge $(i,j)\in \mE$ if and only if $(i,j)\in \mE^{(\omega)}$ for some $\omega \in \Omega$.
\end{defn}

We now provide a definition of factorization of a set of stochastic processes.
\begin{defn}
[Conditional Independence--Frequency Domain (CIFD)]
\label{def:condi_inde_freq}
Consider SPs $\X_1$, $\X_2$, and $\X_3$. Then $\X_1$ is said to be conditionally independent of $\X_2$ given $\X_3$ if and only if for almost all  $\omega \in \Omega$,
\begin{align*}
    \mP\left( \bigcap_{i=1}^2 \left\{\widehat{X}_i(\omega) \in B_{i}\right\} \Bigm\vert \hatt{X}_3(\omega) \in B_{3} \right)=\prod_{i=1}^2\mP\left(  \widehat{X}_i(\omega) \in B_{i}\Bigm\vert \hatt{X}_3(\omega) \in B_{3} \right),
\end{align*}
where $B_i \in \mathcal{B}^2$.
\end{defn}
 Using this definition, one can define factorization of SPs according to graphs, which helps in performing do-calculus and causal identification.
\begin{defn}[Factorization of stochastic process]
\label{def:factorization_SP}
A set $V$ of stochastic processes on $(\Theta,\mF)$ is said to factorize according to $G=(V,\mE)$ if and only if 
\begin{align}
\label{eq:factorization_dist_graph}
\mP\left(\hatt{X}_1(\omega),\dots,\hatt{X}_n(\omega)\right)=\prod_{i\in V} \mP\left(\hatt{X}_i(\omega) \mid \hatt{X}_{Pa_i}(\omega) \right) ~ \text{ for almost all } \omega \in \Omega,
\end{align}
where $Pa_i$ denotes the parents of $i$ in $G$,
\end{defn}

\section{Main Results}
\label{sec:main_results}
\subsection{Computational Advantages of FFT based Wiener filter}

The first contribution of the article delineates the computational complexity when the multivariate Wiener filters are obtained via time-domain projections in comparison to the computational complexity when the Wiener filter coefficients are determined via projection in the frequency domain. In Section~\ref{subsec:computation advantage} we show that the asymptotic computational complexity of determining the Wiener filter coefficient, $W_i$, using the time-domain approach (see equation~\eqref{eq:Wiener_filter_time}) is $O(Tn^2N^2)$ whereas when the Wiener filter is obtained via projections in the frequency domain (see equation~\eqref{eq:Wiener_filter}) it is $O(Tn^2\log N)$; here each time series has $T$ samples, $N$ is the largest time lag, and $n$ is the number of time-series. 
 

\subsection{Realizing the Essential Graph of SPs; using Phase Properties of Frequency Dependent Wiener Filtering}
Mirroring the approach in the PC algorithm \cite{kalisch2007estimating}, we can apply the following approach to reconstruct the essential graph of dynamically related SPs. In the vanilla PC algorithm for reconstructing $G=(V,\mE)$, the first step is to find the topology (also called skeleton) of the graph, $top(G)$ by performing, for every pair $i,j \in V$, conditional independence test on $i$ and $j$ conditioned on $C \subseteq V \setminus \{i,j\}$. If $i$ and $j$ are conditionally independent given $C_{ij}$ for some $C_{ij} \subset V \setminus\{i,j\}$, then $(i,j) \notin top(G)$. Notice that $C_{ij}$ is initialized by setting $C_{ij}=\{\}$, followed by  the nodes with $|C_{ij}|=1$, followed by  $|C_{ij}|=2$, and so on in the increasing order of size. Once the skeleton is retrieved, check for every $(i,j) \in top(G)$ and every $k$ that is a common neighbor of $i$ and $j$, if $k \in C_{ij}$. If $k \notin C_{ij}$, then $k$ is a collider of the form $i\rightarrow k \xleftarrow{} j$. By repeating this process for every nonadjacent pairs $i,j$ and their common neighbors $k$, we can identify all the colliders in $G$. As shown in \cite{materassi2013reconstruction}, multivariate Wiener filter can be employed to determine the CI structure in LDIMs. Let $G=(V,\mE)$ be the graph representation of an LDIM \eqref{eq:LDIM}. Then, $(i,j) \in top(G)$ if and only if $W_{i\cdot [j,C]}[j]=0$. That is, if it holds that, in the projection of $X_i$ on $X_{[j,C]}$ by \eqref{eq:Wiener_filter}, the coefficient of $X_j$ is zero ($i$ and $j$ are Wiener separated by $C$). Then, we can employ the same steps in the vanilla PC algorithm, where conditional independence is tested using Wiener separation, to reconstruct the essential graph of $G$. 
The algorithm is called Wiener-PC (W-PC) algorithm here. We show in  Section \ref{subsec:computation advantage} that the computational complexity of W-PC is $O(RNn^{q+3} \log N)$, where $q$ is the maximum neighborhood size. 

As shown in Lemma \ref{lem:Wiener_imaginary}, in a large class of applications \cite{doddi2019topology}, support of the imaginary part of the frequency dependent Wiener filter retrieves $top(G)$ exactly. Thus, by analyzing the real and imaginary part of $W_i$ separately, one can extract the strict spouse edges. This information can in turn be employed to identify the colliders in $G$ in an efficient way (see Section \ref{subsec:computation advantage}). We call this algorithm Wiener-Phase here. The worst case asymptotic computational complexity of Wiener-Phase algorithm is $O(n^3(RN\log N+q^2))$ (see Section \ref{subsec:computation advantage}), which is advantageous, especially in highly connected graphs (with large $q$). We demonstrate the Wiener-phase algorithm in the simulations section.

\subsection{Causal Inferences on Set of Stochastic Processes via Analysis of Dependencies at Specified Frequencies}

Consider a set of SPs $\X$ on $\pspace$. Let $A,B,C \subset V$ be disjoint sets. Then, as shown in Definition \ref{def:condi_inde_freq}, $A$ is independent of $B$ given $C$, denoted $A \independent B \mid C$, if and only if 
$A \independent^{(\omega)} B \mid C$ for every $\omega \in \Omega$. The following lemma follows immediately.

\begin{lemma}
\label{lemma:d_sep_implies_inde}
Consider a set $V$ of stochastic process $\X$ that factorizes according to a directed graph $G=(V,\mE)$. Let $A,B,C \subset V$ be disjoint sets.  If $\dsep{G^{(\omega)}}{A,C,B}$ then $A \independent^{(\omega)} B \mid C$ and if $\dsep{G}{A,C,B}$, then $A \independent B \mid C$. Further, if the distribution is faithful, then the converse also holds.
\end{lemma}

The following lemma establishes a criterion which allows inference of conditional independence of stochastic processes $\{X_i\}_{i=1}^n$ from conditional independence of  $\{\hatt{X}_i(\omega)\}_{i=1}^n$ obtained by stochastic processes component at a  specific frequency $\omega$.
\begin{lemma}
\label{lem:indep_G_implies_ind_omega}
Consider a set $V$ of stochastic process  that factorizes according $G^{(\omega)}=(V,\mE^{(\omega)})$ at $\omega \in \Omega$. Let $G=(V,\mE)$ be such that $(i,j)\in \mE$ if and only if $(i,j)\in \mE^{(\omega)}$ for some $\omega \in \Omega$. Let $X,Y,Z \subset V$ be disjoint sets. 
If $Z$ d-separates $X$ and $Y$ in $G$, then $Z$ d-separates $X$ and $Y$ in $G^{(\omega)}$.
\end{lemma}
\textbf{Proof:} See Appendix \ref{app:indep_G_implies_ind_omega}.

Much stronger results follow for LDIMs.
The following result  for LDIMs establishes the equivalence of $d-separation$ on the graph $G$ according to which the stochastic processes factor (according to Definition~\ref{def:factorization_SP}) and $d-separation$ on the graph $G^{(\omega)}$ which factors the LDIM at a specific frequency $\omega\in \Omega.$
\begin{lemma}
\label{lem:dsep_G_implies_inde_omega_LDIM}
Consider a set $V$  of stochastic process that is described by the LDIM \eqref{eq:LDIM} (here ${\E}_i(k) \independent {\E}_j(l)$ for $i\neq j$ and $k,l \in \mbZ$). Moreover, suppose the set $V$ of stochastic processes   factorizes according to directed graph $G=(V,\mE)$ and at a specific frequency $\omega\in \Omega$ factorizes according to a directed graph $G^{(\omega)}=(V,\mE^{(\omega)})$. Let $X,Y,Z \subset V$ be disjoint sets. Then $Z$ d-separates $X$ and $Y$ in $G$ if and only if $Z$ d-separates $X$ and $Y$ in $G^{(\omega)}$, for almost all $\omega \in \Omega$.
\end{lemma}
\textbf{Proof:} See Appendix \ref{app:dsep_G_implies_inde_omega_LDIM}.

\begin{lemma}
\label{lem:inde_G_implies_inde_omega_LDIM}
Consider a set $V$  of stochastic process that is described by the LDIM \eqref{eq:LDIM} (here ${\E}_i(k) \independent {\E}_j(l)$ for $i\neq j$ and $k,l \in \mbZ$). Moreover, suppose the set $V$ of stochastic processes   factorizes according to directed graph $G=(V,\mE)$ and at a specific frequency $\omega\in \Omega$ factorizes according to a directed graph $G^{(\omega)}=(V,\mE^{(\omega)})$. Let $X,Y,Z \subset V$ be disjoint sets. Then $X \independent^{(\omega)} Y \mid Z$ if and only if $X \independent  Y \mid Z$, for almost all $\omega \in \Omega$.
\end{lemma}
\textbf{Proof:} See Appendix \ref{app:inde_G_implies_inde_omega_LDIM}

The CI and graph factorization machinery described here can be used in identifying the causal structure, and in estimating direct and total effect. We provide development of single-door, back-door and front-door criteria for causal effect identification in the supplementary material.

\section{Complexity Analysis of Wiener filter} 
\label{subsec:computation advantage}

\subsection{Computational Complexity of Estimating Wiener Filter} 

In the time-domain, consider projection of $T$ samples $\{X_i(t)\}_{t=1}^T$ to past $N$ values (including present) of $X_j(t)$, for every $t =1,\dots,T$. The regression coefficients in the projection of $X_i(t)$ on the rest are $\{h_{ij}(t): 0 \leq k \leq N-1, 1 \leq j\leq n, i\neq j\}$, i.e., $(n-1)N$ variables. The complexity of computing the least square for node $i$ using \eqref{eq:Wiener_filter_time} is $O(T N^2(n-1)^2)\approx O(T N^2n^2)$. Repeating this process $n$ times, the total complexity for the projection of $n$ time-series is $O(T N^2n^3)$.

If we compute FFT for a window size of $N$ samples without overlap, then $N\log N$ computations are required to obtain FFT. Computing the Wiener co-efficient for a given $\omega$ (here the number of samples is $T/N$ since we computed DFT with $N$ samples) using the regression \eqref{eq:Wiener_filter} takes $O(\frac{Tn^2}{N})$ computations, thus a total of $O(Tn^2 \log N)$ computations are required to compute the Wiener coefficient for a single frequency, in contrast to $O(Tn^2N^2)$ in time-domain. Repeating this process $n$ times give an overall complexity of $O(Tn^3 \log N)$ for our approach.

\subsubsection{Wiener meets Peter-Clark}
\label{subsubsec:W-PC}

As shown in \cite{materassi2013reconstruction}, Wiener filtering can identify the CI structure in linear dynamical models. Thus, combining it with PC algorithm gives W-PC algorithm. As shown before, computing $W_i$ using FFT takes a complexity of $O(RNn^3\log N)$. As shown in \cite{kalisch2007estimating}, PC algorithm requires $O(n^q)$ tests, thus giving an overall complexity of $O(RNn^{q+3}\log N)$, where $q$ is the neighborhood size. 

As shown in Section \ref{sec:main_results}, Lemma \ref{lem:Wiener_kin_graph} retrieves the Markov blanket structure in $G$ and Lemma \ref{lem:Wiener_imaginary} returns the skeleton of $G$. Combining both, we propose Wiener-phase algorithm in the supplementary material to identify the colliders and thus the equivalence class of the graphs efficiently. $\mK$ is the set of kin edges, $\mS$ is the skeleton, and $\mSP$ is the strict spouse edges. $\mC_{ij}:=\{k:\{i,k\}\in \mS, \{j,k\}\in \mS, \text{ and } \{i,j\}\in \mSP\}$ is the set of potential colliders formed by $i$ and $j$. For any $\{i,j\} \in \mSP$, if $|\mC_{ij}|=1$, then $\mC_{ij}$ is a collider. If $|\mC_{ij}|>1$, then for every $c \in \mC_{ij}$, we can compute $W_{i\cdot [j, c]}$. If $W_{i\cdot [j, c]} \neq 0$, then $c$ is a collider since $\{i,j\}  \notin \mS$. The complexity of computing $\mK$ and $\mS$ is $O(n^3RN\log N)$. Step $4$ is repeated $O(nq)$ times and step 4 (a) which checks for potential colliders among the common neighbors of $i$ and $j$ takes $O(n^2q)$ operations. 4(b) and 4(c) take $O(n)$. Thus the complexity in computing step 4 is $O(n^3q^2)$, giving the overall complexity of $O(n^3(RN\log N+q^2))$. Notice that it is sufficient to compute Algorithm \ref{alg:Wiener_phase} for $O(1)$ number of $\omega$. Thus, the final complexity remains the same.


\subsection{Sample Complexity Analysis}
\label{subsec:sample_complexity}

In practice, due to finite time effects, the Wiener coefficients estimated can be different. Using the concentration bounds from \cite{doddi2019topology} one can obtain a bound on the estimation error of $W_{ij.Z}(\omega)=$ $\W_{i\cdot[j,Z]}(\omega):=\Phi_{X_i\X_{[j,z]}}(\omega)\Phi_{\X_{[j,z]}}^{-1}(\omega)$. 
\begin{theorem}
\label{Thm:Wiener_bound}
Consider a linear dynamical system governed by \eqref{eq:LDIM}. Suppose that the auto-correlation function $R_x(k):=\mathbb{E}\{X(n)X^T(n+k)\}$ satisfies exponential decay, $\| R_x(k) \|_2 \leq C \delta^{-|k|}$ and that there exists $M$ such that $\frac{1}{M} \leq \lambda_{min}(\Phi_\X)\leq \lambda_{max}(\Phi_\X)\leq M$. Then for any $0<\epsilon$ and $L \geq \log_\delta\left( \frac{(1-\delta) \epsilon}{2C} \right)$,

{\footnotesize\begin{align}
\mP\left( \|W_i-\hatt{W}_i\|>\epsilon \right)    \leq n^2 \exp\left( -(T-L) \min\left\{ \frac{81\epsilon^2}{3200c_1^2M^{16}(2L+1)^2 n^2C^2 },\frac{9\epsilon}{80c_1M^4(2L+1) nC } \right\}\right)
\end{align}}

\end{theorem}
The proof of Theorem \ref{Thm:Wiener_bound} is provided in the supplementary material. For small enough $\epsilon$, the first term is the dominant one. Then $\mP\left( \|W_i-\hatt{W}_i\|>\epsilon \right)<\delta$  
if the number of samples,  
$ T\gtrapprox O\left(\frac{M^{16}L^2 n^2}{\epsilon^2}\ln\left(\frac{n^2}{\delta}\right)\right)$.



\section{Limitations and Future Works}

Frequency-domain independence are defined primarily for linear time-invariant models. It is unclear whether these notions of independence can explain non-linear models. FDD based notion fails for infinite convolution, that is, when the support of $\{H_{ij}(t):t\in I\}$ is not finite. Sample complexity of dynamical causal inference is generally higher than the static counterpart. It is not clear if the sample complexity of frequency-domain approach is higher than the time-domain counterparts in dynamical setup; this is a future study. Employing efficient loss functions instead of vanilla least square loss in FFT domain might improve the computational and the sample complexities, and therefore  require further scrutiny.

\section{Simulation Results}

\begin{figure}
\centering
\begin{subfigure}{0.23\textwidth}
\centering
\includegraphics[trim=50 340 530 35,clip, width=0.9\textwidth]{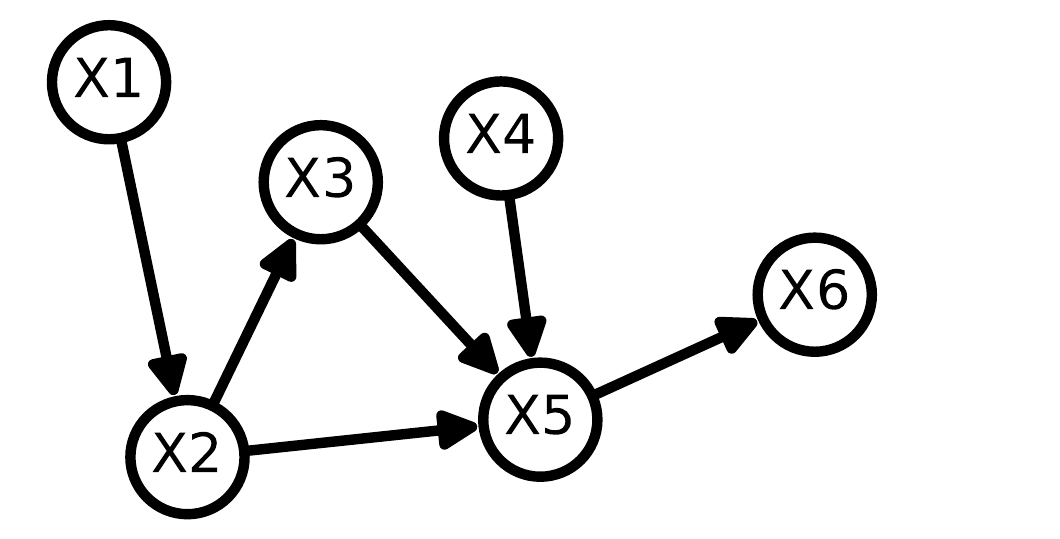} 
\caption{ $G=(V,\mE)$}
\label{fig:causal_graph}
\end{subfigure}
\begin{subfigure}{0.23\textwidth}
\centering
\includegraphics[trim=570 340 60 30,clip, width=0.9\textwidth]{Wiener_phase_Graphs.pdf} 
\label{fig:Kin_graph_Winer}
\caption{Estimated kin-graph}
\end{subfigure}
   \begin{subfigure}{0.23\textwidth}
    \includegraphics[trim=40 20 600 325,clip, width=0.9\textwidth]{Wiener_phase_Graphs.pdf} 
        \caption{Estimated CPDAG}
    \end{subfigure}
        \begin{subfigure}{0.23\textwidth}
        \includegraphics[trim=60 320 600 35,clip, width=0.9\textwidth]{Wiener_phase_Graphs.pdf}
        \caption{Estimated graph}
    \end{subfigure}
    \caption{{\small (a) shows the true causal generative graph; (b) shows the estimated kin-graph from $T=10,000$ samples with $N=64$ using $\ell_\infty(\Omega_N)$ norm on $W_i[j]$; (c) is the CPDAG estimated using Algorithm \ref{alg:Wiener_phase}. The bi-directed edges indicate undirected edges in (c); (d) shows the graph estimated by performing the intervention at node $2$.} \vspace{0pt}}.
\label{fig:simulation_networks}
\end{figure}

The simulations are performed on a synthetic dataset. MacbookPro with M1-pro chip having 8 performance cores (up to 3.2GHz) and 2 efficiency cores (up to 2.6GHz) is used for the simulation. The data is generated according to an LDIM \ref{eq:convolution_model} with the causal graph shown in Figure \ref{fig:simulation_networks}(a). The total number of samples in the time-series data is $T=10,000$ and $N=64$ point FFT is computed. Figure \ref{fig:simulation_networks}(b) shows the estimated kin-graph from the magnitude of $\|W_i[j]\|_{\ell_\infty(\Omega_N)}$, where $W_i$ is computed using \eqref{eq:Wiener_filter}.

\begin{figure}[htb!]
    
    \centering
    \begin{subfigure}{0.45\textwidth}
        \includegraphics[trim=400 420 220 0,clip, width=\textwidth]{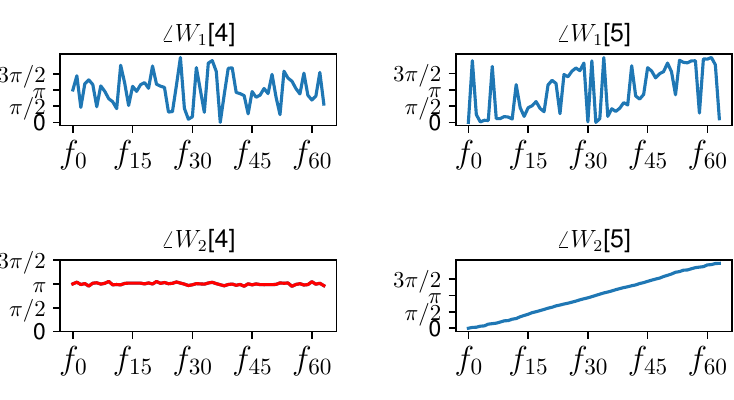} 
        \caption{Phase response plot}
    \end{subfigure}\hfill
    \begin{subfigure}{0.45\textwidth}
        \centering
        \includegraphics[trim=20 20 245 20,clip,width=.85\textwidth]{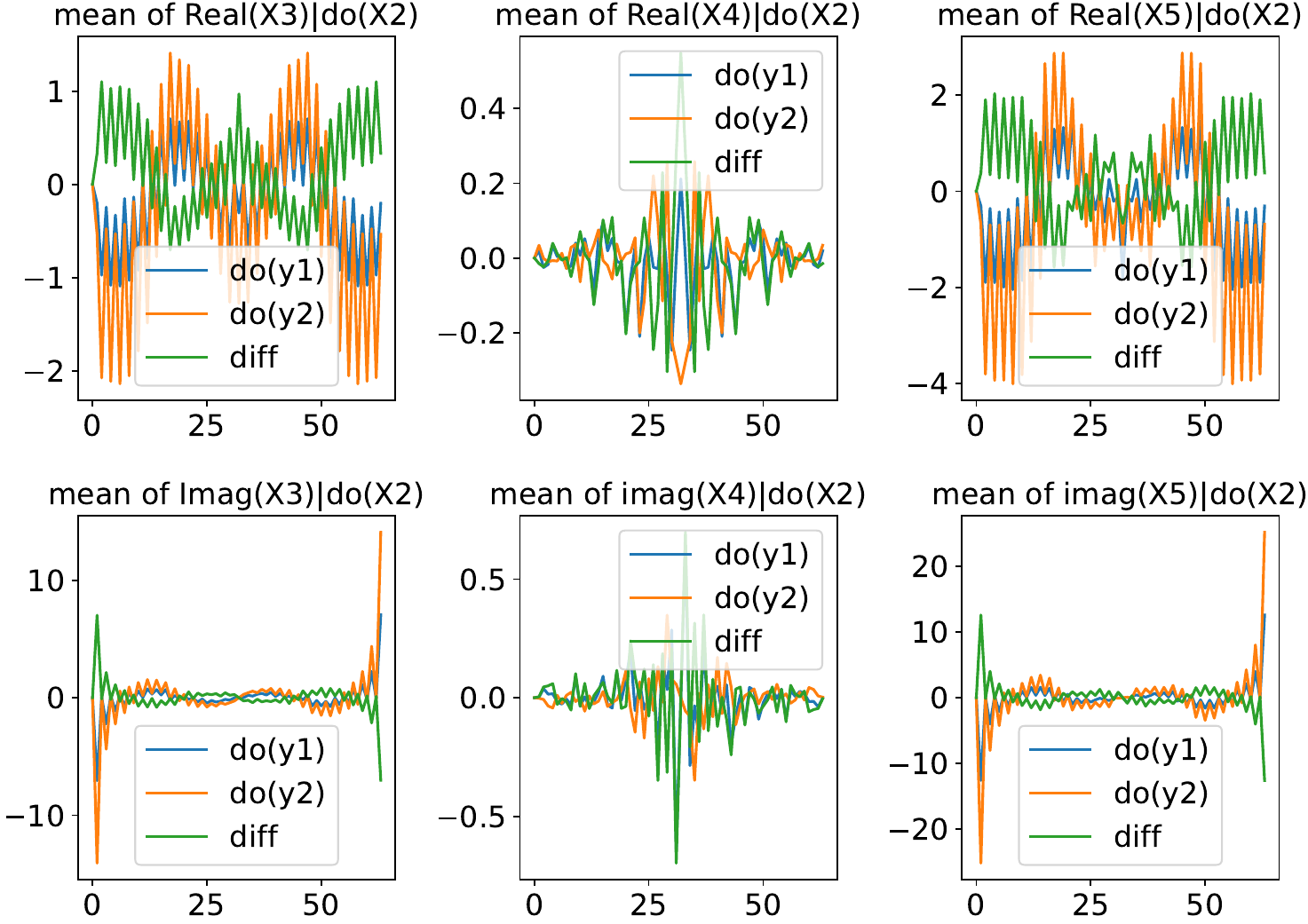} 
        \caption{Estimated conditional distribution}
    \end{subfigure}
    \caption{(a) shows the phase response of $W_i[j]$ for $i=1,2$ and $j=4,5$. (b) shows the plot of real and imaginary parts of the mean vs $\omega$ of the estimated conditional distribution, when conditioned on $do(X_2=y_1)$ and $do(X_2=y_2)$.}
    \label{fig:plots}
\end{figure}

We applied the Wiener-Phase Algorithm to compute the CPDAG in Fig \ref{fig:simulation_networks}(c). Figure \ref{fig:plots} shows the phase response of Wiener filter, $W_i[j]$ for $i=1,2$ and $j=4,5$ versus $\omega$. The phase response plot of $W_2[4]$ is almost constant at $\pi$ (average =3.1 and standard deviation=0.1), whereas the phase response of the remaining pairs in the figure shows a larger variation in mean and variance. Thus, the link $X_2-X_4$ is deemed spurious. In order to compute the rest of the directions, we perform intervention at node $2$ with two different values $X_2=y_1$ and $X_2=y_2$, where $y_2=2y_1$. $y_1$ is a repeating sequence with 32 ones and 32 zeros; this particular pattern is used to obtain a nice 64-point FFT (sync function) for $y_1$ in the frequency-domain. Using this intervention, the time-series data generated by the LDIM was used to obtain the corresponding Fourier transform using FFT. The plots show the variation of mean of real and imaginary parts of $\hatt{X}_i(\omega)$ versus $\omega$ under interventions on $X_2$. It is evident from the estimated conditional distributions plots that $\mP(\hatt{X}_i(\omega)|do(X_2=y_2))$ is different from $\mP(\hatt{X}_i(\omega)|do(X_2=y_1))$ for $i=3$ and $5$, and very similar for $i=4$. It can also be observed from the plots of the imaginary part that large values of mean for $\omega_k$ near $k=0$ and $k=64$. Detailed and expanded delineation of the results can be found in the supplementary material.


\section*{Conclusions}
In this article, we studied the causal identification of linear dynamical systems from the frequency-domain perspective. It was shown that, if the Wiener filter is employed in the FFT domain, the computational complexity of identifying the correlation between the time-series can be reduced significantly. Exploiting the phase properties of the Wiener filter, an algorithm was proposed to efficiently identify the skeleton and the colliders in a graph from data, which can be applied in a large class of applications. Two notions of conditional independence were proposed to realize graph factorization and perform intervention-calculus in stochastic process. Back-door and front-door criteria were derived for stochastic processes using the proposed notions. Non-asymptotic concentration bounds were provide for the error in estimating the Wiener coefficients from data. Finally, theoretical results were verified with simulations.


{
\small



}

\newpage

\appendix

\section*{Appendix}
\textbf{Table of Contents}
\begin{itemize}
\item Wiener-Phase algorithm -- Appendix \ref{app:Wiener-Phase}
\item Simulation -- Appendix \ref{app:simulation}
\item Theorem \ref{thm:convergence} (convergence of FFT) -- Appendix \ref{app:convergence}
\item Proof of Lemma \ref{lem:indep_G_implies_ind_omega} (d-separation in $G$ implies d-separation in $G^{(\omega)}$) -- Appendix \ref{app:indep_G_implies_ind_omega}
\item Proof of Lemma \ref{lem:dsep_G_implies_inde_omega_LDIM} -- Appendix \ref{app:dsep_G_implies_inde_omega_LDIM}
\item Proof of Lemma \ref{lem:inde_G_implies_inde_omega_LDIM} -- Appendix \ref{app:inde_G_implies_inde_omega_LDIM}
\item Theorems on Single-door, back-door and front-door -- Appendix \ref{app:do_calculus}
\item Proof of single-door -- Appendix \ref{app:single_door}
\item Proof of Theorem \ref{app:backdoor} (Back-door) -- Appendix \ref{app:back-door}
\item Proof of Theorem \ref{Thm:Wiener_bound} -- Appendix \ref{app:Wiener_bound}
\item Primer on Wiener filter -- Appendix \ref{app:Wienerproj_timeseries}
\item Primer of stochastic processes and FDD based conditional independence -- Appendix \ref{app:SP}
\end{itemize}

\section{Wiener-Phase Algorithm}
\label{app:Wiener-Phase}
In here, we provide the Wiener-Phase which is described in Algorithm~\ref{alg:Wiener_phase}. As shown in Lemma~\ref{lem:Wiener_imaginary}, in a large class of applications \cite{doddi2019topology}, support of the imaginary part of the frequency dependent Wiener filter retrieves $top(G)$ exactly. Further, as shown in Lemma~\ref{lem:Wiener_kin_graph}, support of $W_i$ retrieves the Markov blanket structure in $G$. Combining both, we can obtain the Wiener-phase algorithm below to identify the colliders and thus the equivalence class of the graphs efficiently. $\mK$ is the set of kin edges obtained using Lemma \ref{lem:Wiener_kin_graph} and $\mS$ is the skeleton obtained from Lemma \ref{lem:Wiener_imaginary}. Consider any edge $(i,j)$ that belongs to $\mK$ but does not belong to $\mS$. Then $i$ and $j$ share a common child without a direct link between them (strict spouses) in $G$. Following this procedure, we can construct all the strict spouses. Let this set be $\mSP$. The complexity of computing $\mK$ and $\mS$ is $O(n^3RN\log N)$

Now consider the skeleton $\mS$. In Step 4, we identify the common child between the strict spouses $i,j$ as follows. For any $i,j \in \mSP$, let $\mC_{ij}:=\{k:\{i,k\}\in \mS, \{j,k\}\in \mS\}$, which is a set that contains all nodes $c$ which has a link to both $i$ and $j$ in the skeleton. Note that the set $\mC_{ij}$ can contain non-colliders, which will be eliminated using the following steps. For any $\{i,j\} \in \mSP$, if $|\mC_{ij}|=1$, then $c\in \mC_{ij}$ is a collider (because the link between $i$ and $j$ in $\mK$ is formed by at least one collider). If $|\mC_{ij}|>1$, then for every $c \in \mC_{ij}$, we can compute $W_{i\cdot [j, c]}$. If $W_{i\cdot [j, c]}[j] \neq 0$, then $c$ is a collider since $\{i,j\}  \notin \mS$. The complexity of computing $\mK$ and $\mS$ is $O(n^3RN\log N)$. Step $4$ is repeated $O(nq)$ times and Step 4(a) which checks for potential colliders among the common neighbors of $i$ and $j$ takes $O(n^2q)$ operations. Steps 4(b) and 4(c) take $O(n)$. Thus the complexity in computing Step 4 is $O(n^3q^2)$, giving the overall complexity of $O(n^3(RN\log N+q^2))$. Notice that it is sufficient to compute Algorithm \ref{alg:Wiener_phase} for $O(1)$ number of $\omega$. Thus, the final complexity remains the same.

\begin{algorithm}[htb]
\caption{Wiener-Phase algorithm}
\noindent \hspace{0cm}\textbf{Input: } Data $\mX(\omega)$, $\omega \in [0,2\pi]$\\
\hspace{0cm}\textbf{Output:} $\hatt{G}$ 
\label{alg:Wiener_phase}
\begin{enumerate}
    \item Initialize the ordering, $\mS\xleftarrow{}$()
    \item For $i=1,\dots,n$ 
    \begin{enumerate}
        \item Compute $W_i(\omega)$ using \eqref{eq:Wiener_filter}  
        \item for $j=1,\dots,n$
        \begin{enumerate}
            \item If $|W_i[j]|>\tau$, then $\mK \xleftarrow{} \mK\cup \{ i,j\}$
            \item If $|\Im\{W_i[j]\}|>\tau$, then $\mS \xleftarrow{} \mS\cup \{ i,j\}$
        \end{enumerate}
    \end{enumerate}
    \item compute $\mSP:=\mK\setminus\mS$
    \item for $\{i,j\}\in \mSP$
    \begin{enumerate}
        \item for $k=1,\dots,n$
        \begin{enumerate}
            \item if $\{i,k\} \in \mS$ and $\{j,k\} \in \mS$
            \begin{itemize}
                \item $C_{ij} \xleftarrow{}C_{ij} \cup k$ 
            \end{itemize}
        \end{enumerate}
        \item If $|C_{ij}|= 1$: $Col\xleftarrow{} Col \cup k$
        \item Else: for $c \in C_{ij}$ compute $W_{i\cdot [j,c]}$
        \begin{enumerate}
            \item If $|W_{i\cdot [j,c]}[j]|>\tau$: $Col\xleftarrow{} Col \cup c$
        \end{enumerate}
    \end{enumerate}
    \item Return $\hatt{G}$
\end{enumerate}
\end{algorithm}
\newpage

\section{Simulation}
\label{app:simulation}
\begin{figure}[htb!]
    \centering
    \begin{subfigure}{.8\textwidth}
        \includegraphics[trim=0 0 0 0,clip, width=\textwidth]{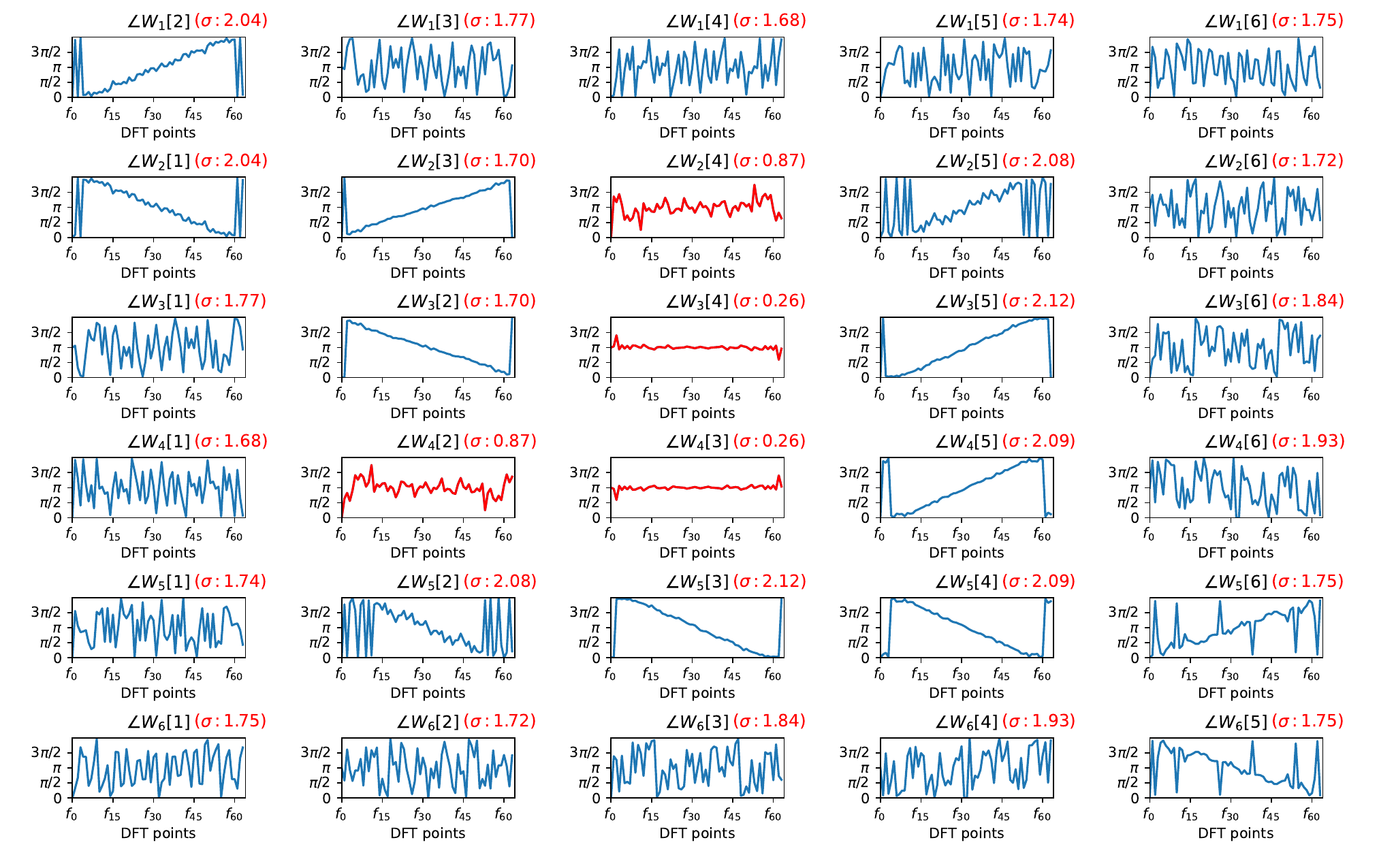} 
        \caption{Phase plot of the Wiener coefficients, $W_i[j]$, computed using equation \eqref{eq:Wiener_filter}; data with 10,000 samples is generated using \eqref{eq:AR_generator}. Red plots show the phase plot, $\angle W_i[j]$, between the strict spouses $i$ and $j$. $\sigma$ is the standard deviation of the phase along the frequency. The spurious links $(i,j)=(2,4)$ and $(i,j)=(3,4)$ have $\sigma<0.9$ whereas the remaining pair of nodes have $\sigma>1.6$.}
    \end{subfigure}\hfill
    \begin{subfigure}{.8\textwidth}
        \centering
        \includegraphics[trim=0 0 0 0,clip,width=\textwidth]{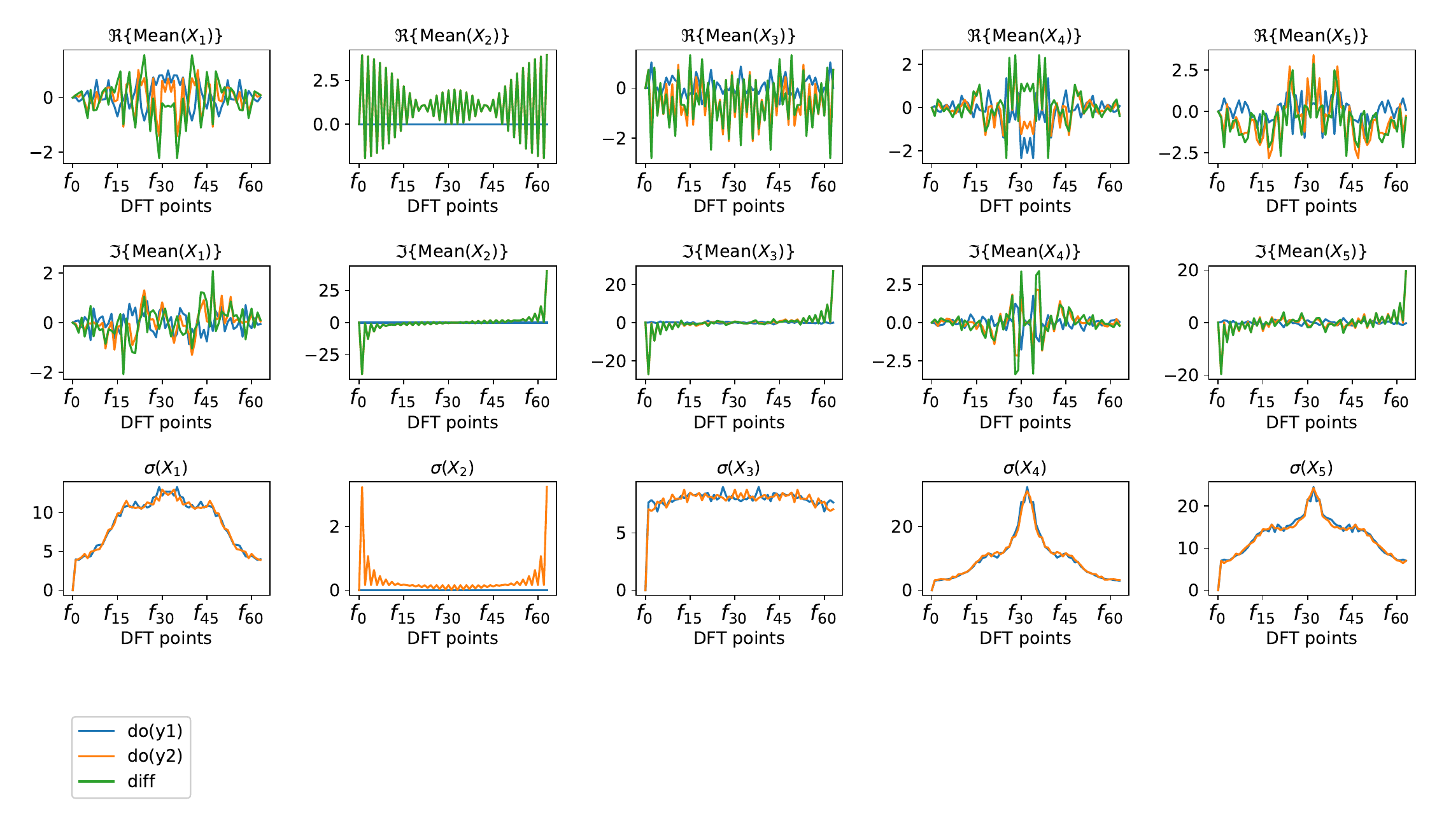} 
        \caption{Estimated conditional distribution when intervened at node $2$ for two interventions, $do(X_2=y_1)$ and $do(X_2=y_2)$, where $y_1$ is all zero sequence and $y_2$ is a repeating sequence of $31$ zeros and $33$ twos. The first row shows the real part of the mean, the middle row is the imaginary part of the mean, and the bottom row is the standard deviation of the estimated distribution. The plots show that the mean of $X_i$, especially the mean of the imaginary part, follows the pattern of $X_2$ when there is a directed path to $i$ from $2$ ($i=3,5$). The mean of $X_1$ and $X_4$ are random, and does not follow the mean of $X_2$, implying that there is no directed path to $1$ and $4$ from $2$.}
    \end{subfigure}
    \caption{(a) shows the phase response of $W_i[j]$, which is the coefficient in the multivariate estimation of the $i^{th}$ time-series corresponding to the $j^{th}$ time-series using \eqref{eq:Wiener_filter}. (b) shows the statistics of $\hX_i(2\pi f)$ vs DFT points. }
    \label{fig:plots_conti}
\end{figure}

The synthetic data is generated using AR model, with the non-zero entries that respect the graph structure shown in Figure  \ref{fig:simulation_networks}(a), 
{\footnotesize\begin{align}
\label{eq:AR_generator}
    \X_i(t)+a_i(1)\X_i(t-1)+a_i(2)\X_i(t-2)+a_i(3)\X_i(t-3)=\sum_{j\neq i}b_{ij} \X_j(t-1)+\E_i(t),
\end{align}}
where $\E_i(t)$ is zero mean i.i.d. Gaussian noise with diagonal covariance matrix and $i=1,\dots,6$. The data is generated continuously according to the AR model in \eqref{eq:AR_generator}. That is, for $t=1$ to $10,000$, $\X(t)$ is computed based on \eqref{eq:AR_generator}, which satisfies Assumption \ref{as:imaginary_non-zero}. Hence Lemma \ref{lem:Wiener_imaginary} and Algorithm \ref{alg:Wiener_phase} can be applied to reconstruct the essential graph. Figure \ref{fig:plots_conti}(a) shows the phase response of the Wiener coefficients, computed using \eqref{eq:Wiener_filter} (Step 2(a) in Algorithm \ref{alg:Wiener_phase}) and Figure \ref{fig:plots_conti}(b) shows the estimated conditional distributions on performing intervention at node $2$.

\begin{figure}[htb!]
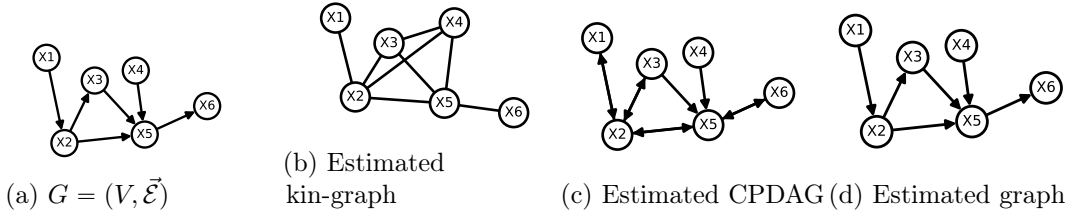

\vspace{0pt}    
\centering
\begin{subfigure}{0.23\textwidth}
\centering
\includegraphics[trim=50 340 530 35,clip, width=0.9\textwidth]{Wiener_phase_Graphs.pdf} 
\caption{ $G=(V,\mE)$}
\label{fig:causal_graph}
\end{subfigure}
\begin{subfigure}{0.23\textwidth}
\centering
\includegraphics[trim=570 340 60 30,clip, width=0.9\textwidth]{Wiener_phase_Graphs.pdf} 
\label{fig:Kin_graph_Winer}
\caption{Estimated kin-graph}
\end{subfigure}
   \begin{subfigure}{0.23\textwidth}
    \includegraphics[trim=40 20 600 325,clip, width=0.9\textwidth]{Wiener_phase_Graphs.pdf} 
        \caption{Estimated CPDAG}
    \end{subfigure}
        \begin{subfigure}{0.23\textwidth}
        \includegraphics[trim=60 320 600 35,clip, width=0.9\textwidth]{Wiener_phase_Graphs.pdf}
        \caption{Estimated graph}
    \end{subfigure}
    \caption{{\small (a) shows the true causal generative graph; (b) shows the estimated kin-graph from $T=10,000$ samples with $N=64$ using $\ell_\infty(\Omega_N)$ norm on $W_i[j]$; (c) is the CPDAG estimated using Algorithm \ref{alg:Wiener_phase}. The bi-directed edges indicate undirected edges in (c); (d) shows the graph estimated by performing intervention at node $2$.} \vspace{0pt}}.
\label{fig:simulation_networks_appendix}
\end{figure}

\newpage
\subsection{Restart and Record Sampling}
\begin{figure}[h!]
    \centering
    \begin{subfigure}{.9\textwidth}
    \includegraphics[trim=0 0 0 0,clip, width=\textwidth]{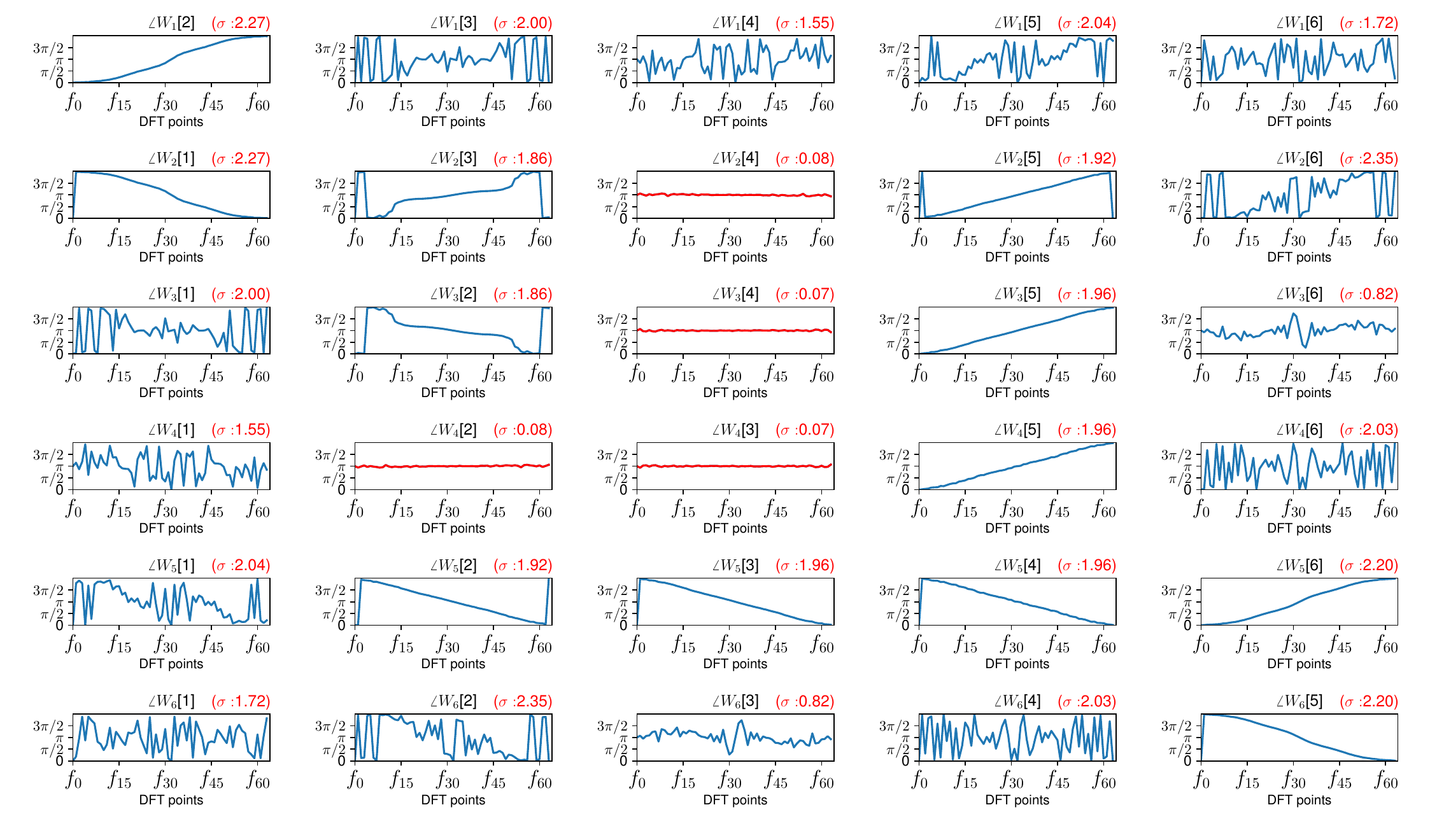} 
        \caption{Phase plot of the Wiener coefficients, $W_i[j]$, computed using equation \eqref{eq:Wiener_filter} for the restart and record data; data with 10,000 trajectories is generated for the AR in \eqref{eq:AR_generator}. Red plots show the phase plot, $\angle W_i[j]$, between the strict spouses $i$ and $j$. $\sigma$ is the standard deviation of the phase along the frequency. The spurious links $(i,j)=(2,4)$ and $(i,j)=(3,4)$ have $\sigma<0.1$ whereas the remaining pairs of nodes have $\sigma>1.5$.}
    \end{subfigure}\hfill
    \begin{subfigure}{.9\textwidth}
        \centering
        \includegraphics[trim=0 0 0 0,clip,width=\textwidth]{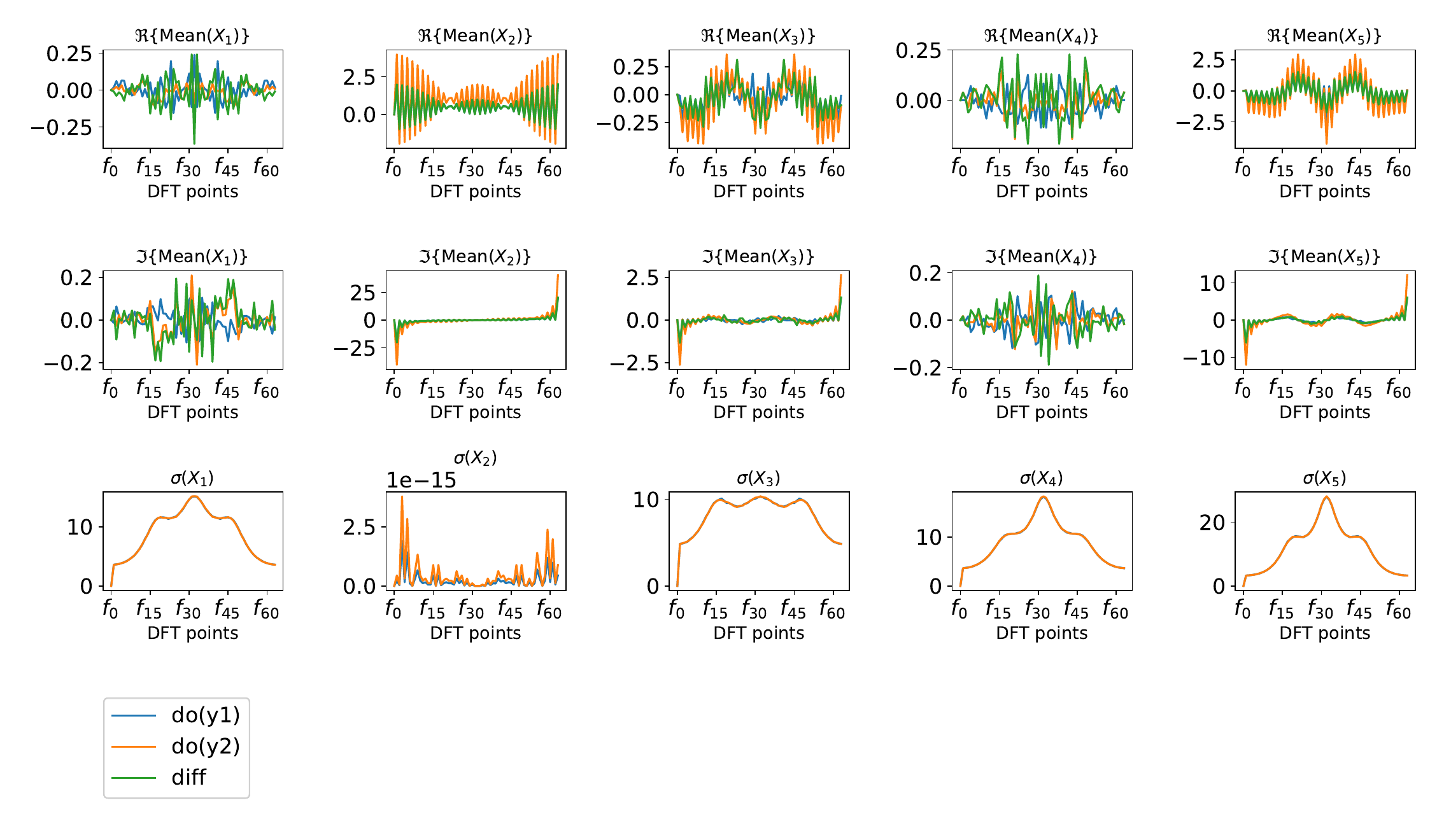} 
        \caption{Estimated conditional distribution when intervened at node $2$ for two interventions, $do(X_2=y_1)$ and $do(X_2=y_2)$, where $y_1$ is all zero sequence and $y_2$ is a repeating sequence of $31$ zeros and $33$ twos.  The plots show that the mean of $X_i$, especially the mean of the imaginary part follows the pattern of $X_2$ when there is a directed path to $i$ from $2$ ($i=3,5$). The mean of $X_1$ and $X_4$ are random, and does not follow the mean of $X_2$, implying that there is no directed path to $1$ and $4$ from $2$.}
    \end{subfigure}
    \caption{Plots obtained for the {\it restart and record data}. (a) shows the phase response of $W_i[j]$, which is the coefficient in the multivariate estimation of the $i^{th}$ time-series corresponding to the $j^{th}$ time-series using \eqref{eq:Wiener_filter}. (b) shows the statistics of $\hX_i(2\pi f)$ vs DFT points. }
    \label{fig:plots_restart_rec}
\end{figure}

In restart and record sampling \cite{doddi2022efficient}, the $i^{th}$ time-series $X_i$ is partitioned into $R$ segments with the $r^{th}$ segment denoted by $X_i^r$. Each segment consists of $N$ samples, for example,  $X_i^r=\{X_i^r(0),\ldots X_i^r(N-1)\}$. The time series $X_i$ is then given by $\{(X_i^r(t))_{t=0}^{N-1}\}_{r=1}^R$. In restart and record sampling, each of the time series is considered independent of the other. Here, following \eqref{eq:AR_generator}, the initial conditions of all $\X_i$ are set to zero. $\E_i(t)$ is generated using i.i.d. Gaussian distribution with mean zero and variance one. Samples of $\X_i(t)$, $i=1,\dots,n$ are realized for $t=0,\dots,N-1$, which are recorded. The procedure is restarted with different realizations for another epoch of $N-1$ samples. Using the $r^{th}$ segment of the  $X_i$ trajectory, given by $X_i^r(0),\dots,X_i^r(N-1)$, the FFT, $\hatt{\X}_i^r(e^\frac{j2\pi k}{N})=\frac{1}{\sqrt{N}} \sum_{n=0}^{N-1} \mathbf{X}_i^r(n) e^{-2\pi kn/N}$ is computed. Figure \ref{fig:plots_restart_rec} shows the plots obtained for the restart and record data. 

\section{Convergence}
\label{app:convergence}

\emph{Notation}: For a continuous function $f$, $\hatt{f}$ denotes the Fourier transform of $f$.
\begin{defn}\cite{grafakos2008classical}
    The total variation of a continuous complex-valued function $f$, defined on $[a,b] \subset \mbR$ is the quantity, $\displaystyle V_a^b(f):=\sup_{p \in \mathcal{P}} \sum_{i=0}^{n_p-1}|f(x_{i+1}-f(x_i)|,$ where $P=\{x_0,\dots,x_{n_p-1}\}$ is a partition and $\mathcal{P}$ is the set of all partitions of $[a,b]$.


    A continuous complex-valued function $f$ is said to be of bounded variation  on a chosen interval $[a,b] \subset \mbR$ if its total variation is finite, i.e.
    if $V_{a}^{b}(f)<+\infty$.
    \end{defn}

\begin{theorem}[Theorem \ref{thm:convergence}]
\label{thm:convergence_appendix}
    For a function $\hatt{f}$ of bounded variation $V$, on $[0,2\pi]$ and $N\geq 1$, the estimation of $f(n)=\int_{0}^{2\pi} \hatt{f}(\omega) e^{j \omega n} d\omega$ given by $f^{(N)}(n)=\frac{1}{N}\sum_{k=0}^{N-1} e^{j2\pi nk/N} \hatt{f}(2 \pi k/N)$ for $|n| \leq \sqrt{N}$ and zero otherwise, satisfies $\|f-f^{(N)}\|_{\ell_\infty(\Omega)} \leq C/\sqrt{N}.$
\end{theorem}


\begin{prop}
\label{prop:int_a_b}
    For $\hatt{f}$ with bounded variation, $V_a^b(\hatt{f})$, on $[a,b]$  and $t\in[a,b]$
    \[\left|\int_a^b \hatt{f}(x) dx-(b-a)\hatt{f}(t)\right| \leq (b-a)V_a^b(\hatt{f}).\]
\end{prop}
\textbf{Proof:} Since $\hatt{f}$ has bounded variation, $M:=\sup_{x\in[a,b]} \hatt{f}(x)$ and $m:=\inf_{x\in[a,b]} \hatt{f}(x)$ are both finite. Thus,
\begin{align*}
    \left|\int_a^b \hatt{f}(x) dx-(b-a)\hatt{f}(t)\right| &\leq  \int_a^b \left|\hatt{f}(x)-\hatt{f}(t) \right|dx\leq (M-m)(b-a) \leq (b-a)V_a^b(\hatt{f}).
\end{align*}
\begin{cor}
    For $\hatt{f}$ a function of bounded variation on $[0,1]$, given by $V_0^1(\hatt{f})$, the error in a Riemann sum approximation, 
    \[\varepsilon_N:=\left|\int_0^1 \hatt{f}(x)dx-\frac{1}{N}\sum_{n=0}^{N-1}\hatt{f}(n/N) \right| \leq V_0^1(\hatt{f})/N. \]
\end{cor}
\textbf{Proof:} Note that \begin{align*}
    \varepsilon_N\leq \sum_{n=0}^{N-1} \left|\int_{n/N}^{n+1/N} \hatt{f}(x)dx-\frac{1}{N}\hatt{f}(n/N) \right|\leq \sum_{n=0}^{N-1} V_{n/N}^{(n+1)/N}(\hatt{f})/N \leq V_0^1(\hatt{f})/N,
\end{align*}
where the second inequality follows from Proposition \ref{prop:int_a_b}.
\begin{prop}[Proposition 3.2.14, \cite{grafakos2008classical}]
\label{prop:Fourier_coeff}
If $\hatt{f}$ is of bounded variation, then
\[{f}(n) \leq \frac{V_0^1(\hatt{f})}{2\pi|n|},~ n \neq 0. \]
\end{prop}
\begin{cor}
\label{cor:Fourier}
    For $n\in \mbN$, and $\varepsilon>0$ fixed. the estimation of the Fourier series ${f}(n)$ given by 
    \[f^{(N)}(n)=\left\{\begin{matrix}
        \sum_{k=0}^{N-1}e^{j2\pi k n/N}\hatt{f}(k/N) & |n| \leq N^{1-\varepsilon}\\
        0 & \text{otherwise}
    \end{matrix} \right.\]
    converges uniformly to ${f}(n)$.
\end{cor}
\textbf{Proof:} For $|n|<N^{1-\varepsilon}$, it follows by Corollary \ref{cor:Fourier} that 
\begin{align*}
    \left| {f}(n)-{f}^{(N)}(n) \right| &\leq \sqrt{2} \frac{\|\hatt{f}\|_\infty 2\pi |n|+V_0^1(\hatt{f})}{N}\\
    & \leq \frac{\sqrt{2} \|\hatt{f}\|_\infty 2\pi}{N^\varepsilon}+ \frac{\sqrt{2} V_0^1(\hatt{f})}{N}.
\end{align*}
Similarly,
for $n\geq N^{1-\varepsilon}$
\begin{align*}
    \left| f(n)-{f}^{(N)}(n) \right| &\leq \frac{V_0^1(\hatt{f})}{2\pi|n|} \leq \frac{V_0^1(\hatt{f})}{2\pi N^{1-\varepsilon}}.
\end{align*}
The proof of Theorem \ref{thm:convergence_appendix} follows by letting $\varepsilon=1/2$. 

\hfill \hfill $\blacksquare$




\section{Proof of Lemmas}
\label{app:indep_G_implies_ind_omega}

\subsection{Proof sketch of Lemma \ref{lem:Wiener_kin_graph}}
Recall that $\Phi_\X^{-1}=(\I-\H)^*\Phi_\E^{-1}(\I-\H)$. Then, for any $i\neq j$, $[\Phi_\X^{-1}]_{ij}=-\H^*_{ji}-\H_{ij}+\sum_{k=1}^n\H_{ki}^*\H_{kj}$. The Lemma follows. Detailed proof is provided in \cite{materassi_tac12}.

\hfill $\blacksquare$
\subsection{Proof of Lemma \ref{lem:indep_G_implies_ind_omega}}
Here, for convenience, we're restating Lemma \ref{lem:indep_G_implies_ind_omega}
\begin{lemma}[Lemma \ref{lem:indep_G_implies_ind_omega}]
Consider a set $V$ of stochastic process  that factorizes according $G^{(\omega)}=(V,\mE^{(\omega)})$ at $\omega \in \Omega$. Let $G=(V,\mE)$ be such that $(i,j)\in \mE$ if and only if $(i,j)\in \mE^{(\omega)}$ for some $\omega \in \Omega$. Let $X,Y,Z \subset V$ be disjoint sets. If $Z$ d-separates $X$ and $Y$ in $G$, then $Z$ d-separates $X$ and $Y$ in $G^{(\omega)}$.
\end{lemma}

\textbf{Proof:} We prove this by contrapositive argument. Suppose $Z$, d-connects $X$ and $Y$ in $G^{(\omega)}$. Then, there exists an $x\in X$ and $y\in Y$ and a path $\pi: x-s_1-\dots s_n-y$ such that $\pi$ is a d-connected path between $x$ and $y$ given $Z$ in $G^{(\omega)}$. 
Then, in $G$, since we are not removing any edge that is present in $G^{(\omega)}$, the d-connected path remains d-connected.

\subsection{Proof of Lemma \ref{lem:dsep_G_implies_inde_omega_LDIM}}
\label{app:dsep_G_implies_inde_omega_LDIM}
\begin{lemma}[Lemma \ref{lem:dsep_G_implies_inde_omega_LDIM}]
Consider a set $V$  of stochastic process that is described by the LDIM \eqref{eq:LDIM} (here ${\E}_i(k) \independent {\E}_j(l)$ for $i\neq j$ and $k,l \in \mbZ$). Moreover, suppose the set $V$ of stochastic processes   factorizes according to directed graph $G=(V,\mE)$ and at a specific frequency $\omega\in \Omega$ factorizes according to a directed graph $G^{(\omega)}=(V,\mE^{(\omega)})$. Let $X,Y,Z \subset V$ be disjoint sets. Then $Z$ d-separates $X$ and $Y$ in $G$ if and only if $Z$ d-separates $X$ and $Y$ in $G^{(\omega)}$, for almost all $\omega \in \Omega$.
\end{lemma}
\textbf{Proof:} The proof follows from Lemma \ref{lem:TF_non-zero}. Recall that $G$ is defined as $G:=\bigcup \limits_{\omega \in \Omega} G^{(\omega)}$. Applying Lemma \ref{lem:TF_non-zero} immediately provides $supp(\H(\omega_1))=supp(\H(\omega_2))$ and $\mG^{(\omega_1)}=\mG^{(\omega_2)}$ for any $\omega_1,\omega_2 \in \Omega$, almost always.  
Thus, $\mG^{(\omega)}=\mG$ almost always in LDIMs. The result follows.

\subsection{Proof of Lemma \ref{lem:inde_G_implies_inde_omega_LDIM}}
\label{app:inde_G_implies_inde_omega_LDIM}
The proof follows by combining Lemma \ref{lem:indep_G_implies_ind_omega} and Lemma  \ref{lem:dsep_G_implies_inde_omega_LDIM}.

\section{Which frequency to choose?}

Most practical systems are finite dimensional with finite dimensional realizations. These admit rational transfer functions.

The following Lemma proves that, in identifying the CI, graph/moral graph, and topology, it is sufficient to work with one frequency selected arbitrarily, in order to get meaningful result, when we have access to large enough data. 
\begin{lemma}
\label{lem:TF_non-zero}
Consider any rational polynomial transfer function $$\H_{ij}(z)=C\frac{(z-a_1)(z-a_2)\dots(z-a_p)}{(z-b_1)(z-b_2)\dots(z-b_q)},~p\leq q,~C\in \mbC.$$ For any distinct $\omega_1, \omega_2 \in \Omega$, $\H_{ij}(e^{j\omega_1}) \neq 0$ if and only if $\H_{ij}(e^{j\omega_2}) \neq 0$ almost surely w.r.t. a continuous probability measure.
\end{lemma}
 \textbf{Proof: }By the fundamental theorem of algebra \cite{IVA}, there exists at most $p$ complex zeros for $\H_{ij}(z)$, if $\H_{ij}(z)$ is not identically zero, which form a set of Lebesgue measure zero. Suppose $\H_{ij}(e^{j\omega_1}) \neq 0$, which implies that $\H_{ij}$ is not identically zero. Thus, $\H_{ij}(e^{j\omega_2}) \neq 0$ almost everywhere.
\hfill $\blacksquare$
 
 \begin{remark}
 Since $\Phi_x^{-1}(e^{j\omega})=(\I-\H(e^{j\omega}))^*\Phi_e^{-1}(e^{j\omega})(\I-\H(e^{j\omega}))$, the same can be said about $\Phi_{\X}(e^{j\omega_1})$ and $\Phi_{\X}(e^{j\omega_2})$ and Wiener coefficients also. Thus, it is sufficient to pick any $\omega \in \Omega$ and apply the algorithms for that frequency.
 \end{remark}

\section{Direct and Total Effect Identification in SPs}
\label{app:do_calculus}

\subsection{Direct Effect Identification}
In the classical causal inference from IID data, some of the popular approaches on parameter identification involve single-door, back-door and front-door criterion \cite{pearl2016causal}. In this article, we extend these results to dynamically related stochastic processes by realizing them in frequency-domain. For the single-door criterion (with cycles) in frequency-domain (for WSS processes), first proved in \cite{Salapaka_signal_selection}, we provide a simpler proof.
\begin{theorem}[Single-door criterion]\
\label{thm:single_door}
Consider a directed graph, $G=(V,\mE)$, that represents an LDIM, defined by \eqref{eq:LDIM}. Let $U,Y \in V$ and $Z\subset V$ with 
$X_Y(\omega)=\alpha(\omega) X_U(\omega)+ \sum_{p \in Pa(Y) \setminus U} H_{Yp}(\omega) X_p(\omega)+E_{Y}(\omega),$ 
for every $\omega\in \Omega$, where $\alpha(\omega)$ is the weight of the link $U\xrightarrow[]{}Y$. Let $G':=G_{\overline{U\rightarrow Y}}$ be the graph obtained by deleting the edge $U\xrightarrow[]{}Y$. Suppose that $U,Y$, and $Z$ satisfy the single door criteria described by: (C1) $Y$ has no descendants in $Z$; (C2) $Z$ d-separates $U$ and $Y$ in $G_{\overline{U\rightarrow Y}}$; (C3) There is no directed cycle involving $Y$. Then the coefficient of $U$ in the projection of $Y$ to $[U,Z]$, denoted, $W_{YU\cdot Z}(\omega)=\alpha(\omega)$.
\end{theorem}

\textbf{Proof:} See Appendix \ref{app:single_door}

Suppose there exists $Pa_i:=Pa(i) \subset V$, $i\in V$ such that the factorization in \eqref{eq:factorization_dist_graph_omega} and \eqref{eq:atomic_intervention_density_split} exist. Then one can perform atomic interventions and the do-calculus on the SPs by performing them at every $\omega \in \Omega$.

\subsection{Total Effect Identification}
One can define the back-door criterion on SPs using CIFD as follows. Notice that a similar result can be obtained using the FDD based independence notion also. The results in this section require the following definitions. A {\it directed path} from node $i$ to node $j$ in the directed graph $G$ is an ordered set of  edges $((\ell_0,\ell_1),(\ell_1,\ell_2),\dots,(\ell_{n-1},\ell_n))$ with $\ell_0=i$, $\ell_n=j$ in  $G=(V,\mE)$, where  $(\ell_i,\ell_{i+1}) \in \mE$.

\begin{defn}[Back-door Criterion] 
\label{def:back-door}
Given an ordered pair of SPs $(W, Y)$ in a DAG $G^{(\omega)}$, a set of SPs $Z$ satisfies the back-door criterion relative to $(W, Y)$ if the following hold:
\begin{itemize}
    \item No node in $Z$ is a descendant of $W$ and
    \item $Z$ blocks (or d-separates) every path between $W$ and $Y$ that contains an arrow into $W$.
\end{itemize}
\end{defn}
\begin{theorem}[Back-door Adjustment]
\label{thm:back-door}
Consider a set of stochastic processes $\{\X_V\}:=\X=[\X_1,\dots,\X_n]$, where $V=\{1,\dots,n\}$. Suppose that $G^{(\omega)}$ is a DAG compatible with $\mP$ with the pdf given by $f^{(\omega)}$. Further, suppose that the processes $W,Y,Z\subset V$ are such that $W$ and $Y$ are disjoint and $Z$ satisfies the back-door criterion (Definition \ref{def:back-door}) relative to $(W,Y)$). Then, the causal effect of $W$ on $Y$ (for notational simplicity $W,Y,Z$ denotes $\X_W,\X_Y,\X_Z$) is identifiable and is given by
$f^{(\omega)}\left(y(\omega) \mid do(W=w) \right)=\int_{}f^{(\omega)}\left(y(\omega) \mid w(\omega), z(\omega) \right)  f^{(\omega)}\left( z(\omega) \right) d \nu(z(\omega))~ \omega \in \Omega.$
\end{theorem}
\textbf{Proof:} See Appendix \ref{app:backdoor}
%

\begin{defn}[Front-door Criterion] 
\label{def:frontdoor}
Given an ordered pair of sets $(W, Y) \subset V \times V$ in a DAG $G^{(\omega)}$, a set of SPs $Z \subset V$ satisfies the front-door criterion relative to $(W, Y)$ if the following hold:
\begin{itemize}
    \item Every directed path from $W$ to $Y$ has a node $z\in Z$.
    \item All back-door paths between $Z$ and $Y$ are blocked by $W$. That is, $W$ d-separates every path between $Z$ and $Y$ which has an arrow into $Z$ and
    \item There are no back-door paths activated between $W$ and $Z$. That is, the empty set d-separates any path between $W$ and $Z$ with arrow into $W$.
\end{itemize}
\end{defn}

\begin{theorem}[Front-door adjustment]
\label{thm:frontdoor}
Consider a set of stochastic processes $\{\X_V\}:=\X=[\X_1,\dots,\X_n]$, where $V=\{1,\dots,n\}$. Suppose that $G^{(\omega)}$ is a DAG compatible with $\mP$ with the pdf given by $f^{(\omega)}$. Further, suppose that the sets $W,Y,Z\subset V$ are such that $W$ and $Y$ are disjoint and $Z$ satisfies the front-door criterion (Definition \ref{def:frontdoor}) relative to $(W,Y)$. Then, the causal effect of $W$ on $Y$ (for notational simplicity $W,Y,Z$ denotes $\X_W,\X_Y,\X_Z$) is identifiable and is given by:

{\footnotesize\begin{align*}
    f^{(\omega)}\left(y(\omega) \mid do(W=w) \right)&=\int_{}f^{(\omega)}\left(z(\omega) \mid w(\omega)\right)\int_{}f^{(\omega)}\left(y(\omega) \mid w'(\omega), z(\omega) \right) f^{(\omega)}\left( z(\omega) \right) d \nu(w'(\omega)) d \nu(z(\omega)).
\end{align*}}
\end{theorem}
\textbf{Proof:} The proof follows similar to the back-door criterion.
\begin{remark}
    One can extend the rules of do-calculus also to the SPs in the same way and is skipped due to space constraints.
\end{remark}

\section{Proof of Theorem \ref{thm:single_door} (Single-door)}
\label{app:single_door}


The following Proposition is useful in proving the succeeding results.

\begin{lemma}[Linearity of Wiener Coefficient]
\label{lemma:linearity_Wiener_coeff}
Let $X,Y, Z_1,\dots,Z_n$ be stochastic processes, and let $\widetilde{Z}:=span(X,Z_1,\dots,Z_n)$ and $\Z=[Z_1,\dots,Z_n]$. Let $W_{YX\cdot \Z}$ be the coefficient of $X$ in projecting $Y$ on to $\widetilde{Z}$. Suppose $Y(\omega)=\sum_{k=1}^m \alpha_k(\omega) Y_k(\omega)$, where $\alpha_1,\dots,\alpha_m$ are transfer functions and $m\in \N$.
Then, 
\begin{align*}
W_{YX \cdot \Z}(\omega) =\sum_{i=1}^m\alpha_i(\omega)W_{Y_iX \cdot \Z}(\omega)
\end{align*}
\end{lemma}
\textbf{Proof: }See Appendix \ref{app:linearity_Wiener_coeff}.

The following lemma, which follows from \cite{Salapaka_signal_selection} shows the relation between d-separation and Wiener coefficients.
{\begin{lemma}
\label{lem:W_zero_d_separation}
Consider a directed graph $G=(V,\mE)$ that represents an LDIM, \eqref{eq:LDIM}. Let $X,Y,Z \subset V$ be disjoint sets. Suppose that $\dsep{G}{X,Z,Y}$. Then  $W_{ij\cdot Z}=0$, for every $i\in X$ and $j\in Y$.
\end{lemma}
\textbf{Proof: }The result follows from Theorem 24 in \cite{Salapaka_signal_selection}.

Now we can prove the single-door criterion.

\subsection{Proof: single-door}
\emph{Notation:} Here, $W^{(L')}_{pU\cdot Z}$ denotes the co-efficient of $\X_U$ in the projection of the time-series $\X_p$ to $\X_{[U,Z]}$. 

Claim 1: If $U$ and $Y$ are d-separated by $Z$ in graph $G'$, then $U$ and any $p \in Pa(Y)\setminus (Z \cup\{U\})$ are also d-separated by $Z$ in $G'$.  

\textbf{Proof:} The claim can be proved by contrapositive argument. Suppose $U$ and $p \in Pa(Y)\setminus \{U\}$ be d-connected by $Z$ in $G'$. Then, there exists a path $PT: \{U,\pi_1,\dots,\pi_n,p\}$ in $G'$ such that 
\begin{enumerate}
    \item If the path is a chain then $\pi_i ,p\notin Z$ (since the path is disconnected otherwise).
    \item If there exists a fork $\pi_j$ in the path, then $\pi_j \notin Z$.
    \item If there are colliders in $(\pi_i)_{i=1}^n$, then either the colliders or their descendants are present in $Z$.
\end{enumerate}
Presence of the path $PT$ implies that there exists a path $PT_Y:=\{U,\pi_1,\dots,\pi_n,p\rightarrow Y\}$, which is obtained by adding $p\rightarrow Y$ to $PT$. 
In the path $PT_Y$, $p$ can be a fork element or a chain element, which does not belong to $Z$. Thus, the path $PT_Y$ remains d-connected given $Z$ in $G'$ 
, which contradicts the d-separation assumption.

Claim 2: For any $p\in Pa(Y)\setminus (Z \cup \{U\} )$ $\dsep{G'}{U,Z,p} \implies \dsep{G}{U,Z,p}$

Suppose $U$ and $p$ are d-connected by $Z$ in $G$. Then, there exists a path $PT:(U,p_1,\dots,p_n,p_{n+1}=p)$ in $G$ such that 
\begin{enumerate}
    \item If $PT$ is a chain, then $p_i,p \notin Z$.
    \item If there exists a fork $p_j$, then $p_j \notin Z$.
    \item If there are colliders in $(p_i)_{i=1}^n$, then for every collider, either the collider or its descendant is present in $Z$.
\end{enumerate}
Notice that the only difference between $G$ and $G'$ is the absence of edge $U\xrightarrow{}Y$ in $G'$.

We devide the proof into two cases
\begin{itemize}
    \item[Case 1:] Suppose the path $PT=U-\pi_1-\pi_2-\cdots \pi_n-p$ has no colliders in the graph $G$. Then it has atmost one fork.
    \begin{enumerate}
        \item Suppose the path $PT$ has one fork at $\pi_k$. Thus the path $PT\equiv U\leftarrow \pi_1\leftarrow \pi_2\leftarrow\cdots \pi_{k-1}\leftarrow \pi_k\rightarrow \pi_{k+1}\rightarrow \cdots\rightarrow p.$ Suppose the subpath $PT_2\equiv \pi_k\rightarrow \pi_{k+1}\rightarrow \cdots\rightarrow p$ has the link $U\rightarrow Y$. Then the path 
       $ \pi_k\rightarrow \pi_{k+1}\rightarrow \cdots\rightarrow U\rightarrow Y\rightarrow  \cdots\rightarrow p\rightarrow Y$ exists in $G$ which involves $Y$ in a directed loop leading to a contradiction. Thus $U\rightarrow Y$ is not present in $PT_2.$ 
       Now suppose the link $Y\leftarrow U$ is present in the subpath $PT_1\equiv U\leftarrow \pi_1\leftarrow \pi_2\leftarrow\cdots \pi_{k-1}\leftarrow \pi_k.$ Suppose $PT_1\equiv U\leftarrow \pi_1\leftarrow \pi_2\leftarrow Y=\pi_{l-1}\leftarrow U=\pi_l\leftarrow \cdots \pi_{k-1}\leftarrow \pi_k$ which implies $Y$ is in a directed loop leading to a contradiction. Thus the path $PT$ does not have a link of the form $U\rightarrow Y$ and remains d-connected by $Z$ in $G'.$
    \item Suppose the path has no forks. As there are no colliders  the path is a chain. (a) Suppose the path, $PT$ is chain from $U$ to $Y$. Suppose there is  the link $U\rightarrow Y$ as  a subpath in $PT$. Then $Y$ will be a part of a directed loop which leads to a contradiction. (b) Suppose the path $PT$ is a chain from $p$ to $U$. Suppose there a link $Y\leftarrow U$ in $PT.$ Again, $Y$ will be part of a directed loop which is a contradiction. Thus the $PT$ remains d-connected by $Z$  in $G'$.
    \end{enumerate} 
\item[Case2:] Suppose the path  $PT$  has $m$ colliders, $c_1\cdots,c_m$  with
$PT\equiv U-\cdots-\rightarrow c_1\leftarrow -\cdots-\rightarrow c_2\leftarrow \cdots -\rightarrow c_m\leftarrow -\cdots p.$ 
Consider the subpath $PT_i\equiv c_i\leftarrow \pi_{1}-\cdots-\pi_q\rightarrow c_{i+1}$  of $PT$ between $c_i$ and $c_{i+1}$ which are colliders. As the path $PT$ is d-connected by $Z$, $c_i$ and $c_{i+1}$ or their descendants are in $Z$. Without loss of generality we assume the path 
$z_\ell=\ell_k\leftarrow \ell_{k-1}\leftarrow \cdots \leftarrow \ell_1\leftarrow c_i\leftarrow \pi_1-\cdots-\pi_q\rightarrow c_{i+1} \rightarrow r_1\rightarrow r_2\cdots \rightarrow r_s=z_r$ where $z_r$ and $z_\ell$ are descendants of $c_{i+1}$ and $c_i$ in $Z$ respectively exists (d-connecting the path). Note that none of the nodes $\ell_j$ or $r_j$ can be $Y$ as no descendant of $Y$ can be in $Z.$ Thus the link $U\rightarrow Y$ cannot be present in the subpath $PT_{i\ell}\equiv z_\ell=\ell_k\leftarrow \ell_{k-1}\leftarrow \cdots \leftarrow \ell_1\leftarrow c_i$ or in $PT_{ir}\equiv c_{i+1} \rightarrow r_1\rightarrow r_2\cdots \rightarrow r_s=z_r.$  

The subpath $PT_i$ has no colliders and is a path between colliders and thus it has one fork say at  $\pi_t$ so that the path $PT_i\equiv c_i\leftarrow \pi_{1}-\cdots- \leftarrow \pi_t\rightarrow-\cdots \rightarrow  c_{i+1}.$ Without loss of generality assume $Y$ appears in the subpath  $ c_i\leftarrow \pi_{1}\leftarrow \cdots \leftarrow \pi_t.$ Then examining the path $PT_{i\ell}$ concatenated with  $c_i\leftarrow \pi_{1}\leftarrow \cdots \leftarrow \pi_t$ results in $Y$ having a descendant in $Z$ which is a contradiction. Thus $Y$ cannot be in $c_i\leftarrow \pi_{1}-\cdots- \leftarrow \pi_t$. Similarly $Y$ cannot be in $\pi_t\rightarrow\cdots \rightarrow  c_{i+1}.$ Thus the link $U\rightarrow Y$ cannot be in $PT_i.$ In conclusion, the link $U\rightarrow Y$ cannot be in  $z_\ell=\ell_k\leftarrow \ell_{k-1}\leftarrow \cdots \leftarrow \ell_1\leftarrow c_i\leftarrow \pi_1-\cdots-\pi_q\rightarrow c_{i+1} \rightarrow r_1\rightarrow r_2\cdots \rightarrow r_s=z_r$ which implies the path $PT_i$ remains d-connected by $Z$ in $G'$.

Thus the $(U,p)$ remain d-connected by $Z$ in $G'$. 

\end{itemize}
    Thus, $\dsep{G'}{U,Z,p} \implies \dsep{G}{U,Z,p}$ and it follows from Lemma \ref{lem:W_zero_d_separation} that $W^{(L)}_{p U\cdot Z}=0$. Notice that if $p \in Z$, then $W^{(L)}_{p U\cdot Z}=0$.
From Linearity of $W$, Lemma \ref{lemma:linearity_Wiener_coeff},
\begin{align*}
W^{(L)}_{Y,U\cdot Z}&= W^{(L)}_{(\alpha U+ \sum_{p \in Pa(Y) \setminus U} a_{Yp} p + \varepsilon_{Y}),x\cdot Z}\\ 
&=\alpha W^{(L)}_{U,U\cdot Z}+\sum_{p \in Pa(Y) \setminus \{x\}} a_{Y,p}W^{(L)}_{p,U\cdot Z}+W^{(L)}_{\varepsilon_{Y},U\cdot Z}\\
&= \alpha + 0.
\end{align*} 
If $Pa(Y)\setminus \{U\}=\emptyset$, then the above equations hold.


\section{Proof of Theorem \ref{thm:back-door}}
\label{app:back-door}
\label{app:backdoor}

\subsection{Atomic interventions} 

{\scriptsize\begin{align}
\label{eq:do_i_definition}
    \mP\left( X_{(1)}(\omega),\dots,X_{(n)}(\omega) \mid do(x_{i} \right)= \left\{ \begin{array}{cc}
        \displaystyle \prod_{k=1, k\neq i}^n\mP\left(X_{k}(\omega) \mid X_{(Pa_k)}(\omega) \right)&  \text{ if } X_{i}(\omega)=x_{i}(\omega){}\\
         0& \text{ otherwise} 
    \end{array} \right.
\end{align}}

\begin{defn}[The Causal Markov Condition]
Suppose that we want to condition on ${X_{i}}(\omega)=x_{i}(\omega)$. Then, the intervened distribution is obtained by removing the factor $\mP\left(X_{i}(\omega) \mid X_{Pa_i}(\omega)\right)$ from the product in \eqref{eq:factorization_dist_graph_omega}. That is, the distribution satisfies \eqref{eq:do_i_definition}. Alternately, $\mP(\cdot)$ can be replaced with the pdf $f(\cdot)$ if $f$ exists.
\end{defn}
We first prove the following auxiliary theorem and lemma.
\begin{theorem}[Causal effect]
\label{thm:Causal_effect}
Let $V:=\{1,\dots,n\}$, $A \subset V$ and let $i \in V$. The causal effect of ${X}_i$ on a set of SPs $X_{A}$ can be identified if $i \cup A \cup Pa_{i}$ are observable and $A \cap (i \cup Pa_i)=\emptyset$. The causal effect of $X_{i}$ on $X_{A}$ is given by

{\footnotesize\begin{align*}
f^{(\omega)}\left( x_{A}(\omega) \mid {x_{i}}(\omega) \right)&=\int f^{(\omega)}\left(x_{A}(\omega) \mid x_{i}(\omega),x_{Pa_i}(\omega)\right)f^{(\omega)}\left( x_{Pa_i}(\omega) \right) d\nu\left(x_{Pa_i}{(\omega)}\right),    
\end{align*}}
if $X_{i}(\omega)=x_{i}(\omega)$ and zero otherwise.
\end{theorem}
\textbf{Proof}: 
The proof follows by Baye's rule and the definition of conditional probability, $\mP(Y \mid Z)=\frac{\mP(Y,Z)}{\mP( Z)}$. If $X_{i}={x_{i}}$, applying \eqref{eq:do_i_definition},
\begin{align*}
    \mP\left( {X_{1}}(\omega),\dots,{X_{n}}(\omega) \mid {{X_{i}}(\omega)} \right)&=\prod_{k=1, k\neq i}^n\mP\left({X_{k}}(\omega) \mid {X_{Pa_k}}(\omega) \right)\\
    & =\frac{\prod_{k=1}^n\mP\left({X_{k}}(\omega) \mid {X_{(Pa_k)}}(\omega) \right)}{\mP\left({X_{i}}(\omega) \mid {X_{Pa_i}}(\omega) \right)}\\
    &=\frac{\mP\left({X_{1}}(\omega),\dots,{X_{n}}(\omega) \right)}{\mP\left({X_{i}}(\omega) \mid {X_{Pa_i}}(\omega) \right)}\\
    &\hspace{0cm}=\mP\left({X_{1}}(\omega),\dots,{X_{n}}(\omega)\mid {X_{i}}(\omega),{X_{Pa_i}}(\omega) \right)\mP\left( {X_{Pa_i}}(\omega) \right).
\end{align*}

Let $U={(V \setminus (\{i\} \cup A)}$ and $U_i:=V \setminus (\{i\} \cup A \cup Pa_i)$. Let $f$ be the joint density function. Then,
\begin{align*}
    f\left( {x_{A}}(\omega) \mid { {x_{i}}}(\omega) \right)&\hspace{0cm}\stackrel{(a)}{=}\int f\left( {x_{V}}(\omega) \mid {x_{i}}(\omega) \right) d\nu(x_{U}(\omega))\\
    &\hspace{0cm}\stackrel{(b)}{=}\int f\left(x_{V} \mid x_{i},x_{Pa_i} \right) f\left( x_{Pa_i}\right) d\nu \left(x_{U}\right)\\
    &\hspace{0cm}\stackrel{(c)}{=}\int_{}~\int_{}f\left(x_V \mid x_i,x_{Pa_i} \right) f\left( x_{Pa_i} \right) d\nu\left(x_{U_i}\right)d\nu\left(x_{Pa_i}\right)\\
    &\hspace{0cm}=\int f\left( x_{Pa_i} \right)\int_{}  f\left(x_V \mid x_i,x_{Pa_i} \right) d\nu\left(x_{U_i}\right)d\nu\left(x_{Pa_i}\right) \\
    &\hspace{0cm}=\int f\left( x_{Pa_i} \right)\int{} \frac{f\left(x_V\right)}{f\left( x_i,x_{Pa_i} \right)} d\nu\left(x_{U_i}\right)d\nu\left(x_{Pa_i}\right) \\
&\hspace{0cm}\stackrel{(d)}{=}\int f\left( x_{Pa_i} \right)\frac{f\left(x_A,x_i,x_{Pa_i}\right)}{f\left( x_i,x_{Pa_i} \right)} d\nu\left(x_{Pa_i}\right)\\
&\hspace{0cm}=\int f\left(x_A \mid x_i,x_{Pa_i}\right)f\left( x_{Pa_i} \right) d\nu\left(x_{Pa_i}\right),
\end{align*}
where $(a)$ follows by marginalization, $(b)$ definition of conditional probability, $(c)$ Fubini's theorem, and
$(d)$ follows by marginalization over $U_i$.
Thus, $f\left( x_{A} \mid {x_{i}} \right)$ can be determined if $X_{A}$, $X_{i}$, and $X_{Pa_i}$ are measured. \hfill $\blacksquare$

The following is an auxiliary Lemma useful in proving back-door criterion. 
\begin{lemma}
\label{lem:back-door_prel}
Consider a set of stochastic processes $\X:=\X_V=\{X_1,\dots,X_n\}$, where $V=\{1,\dots,n\}$. Suppose $G^{(\omega)}$ is a directed graph compatible with $\mP(.)$ at $\omega$. Further, suppose that the processes $W,Y,Z\subset V$ are such that 
\begin{enumerate}
    \item $W\independent^{(\omega)} Z \mid Pa_W$,
    \item $Y\independent^{(\omega)} Pa_W \mid W,Z$
\end{enumerate}
where $Pa_W \subset {V}$ is the set of parents of $W$. Then,
\begin{align*}
    f^{(\omega)}\left(y(\omega) \mid do(W=w) \right)&=\int_{}f^{(\omega)}\left(y \mid w(\omega), z(\omega) \right)f^{(\omega)}\left( \omega \right) d \nu(z(\omega)).
\end{align*}
\end{lemma}
\textbf{Proof: } Let $U=Pa_W$. From Theorem \ref{thm:Causal_effect} (ignoring the index $\omega$),
{
\begin{align*}
    f\left( y \mid do(W=w) \right)&=\int_{} f\left(y \mid w,u \right)f(u) d\nu(u)  \\
    &\stackrel{(a)}{=}\int_{}\int f\left(y \mid w,u,z\right)f(z\mid u,w)f(u) d\nu(z)d\nu(u)  \\
    &\stackrel{(b)}{=}\int \int_{} f\left(y \mid w,z\right)f(z\mid u,w)f(u) d\nu(z)d\nu(u)  \\
    &\stackrel{(c)}{=}\int \int_{} f\left(y \mid w,z\right)f(z\mid u)f(u) d\nu(u)d\nu(z)  \\
    &\stackrel{}{=}\int f\left(y \mid w,z\right) \left[\int_{} f(z\mid u)f(u) d\nu(u)\right]d\nu(z)  \\  &\stackrel{(d)}{=}\int f\left(y \mid w,z\right) f(z)d\nu(z) .
\end{align*}
}
where $(a)$ follows by marginalization over $Z$, $(b)$ $Y \independent U \mid W,Z$, $(c)$ $z \independent w \mid u$, and $(d)$ marginalization. \hfill $\blacksquare$

\subsection{Back-door Criterion}

\begin{defn}[Back-door Criterion] 
\label{app:def:back-door}
Given an ordered pair of SPs $(W, Y)$ in a DAG $G^{(\omega)}$, a set of SPs $Z$ satisfies the back-door criterion relative to $(W, Y)$ if the following hold:
\begin{itemize}
    \item No node in $Z$ is a descendant of $W$ and
    \item $Z$ blocks (or d-separates) every path between $W$ and $Y$ that contains an arrow into $W$.
\end{itemize}
\end{defn}

\textbf{Proof of Theorem \ref{thm:back-door}:} The proof follows from Lemma \ref{lem:back-door_prel}. If $Z$ does not contain descendants of $W$ then $Z \independent W \mid Pa_W$. If $Z$ blocks every back-door path between $Y$ and $W$, then $Y \independent Pa_W \mid W,Z$, given that $Z$ is not a descendent of $W$. The absence of a loop involving $W,~Y$, and $Z$ ensures that $Z$ is not a descendent of $W$.
The result follows. $\blacksquare$

\section{Proof of Theorem \ref{Thm:Wiener_bound}}
\label{app:Wiener_bound}
\begin{lemma}
[Theorem 6, \cite{doddi2019topology}]
\label{lem:PSD_bound}
Consider an linear dynamical system governed by \eqref{eq:LDIM}. Suppose that the autocorrelation function $R_x(k)$ satisfies exponential decay, $\| R_x(k) \|_2 \leq C \delta^{-|k|}$. For any $0<\epsilon_1<\epsilon$ and $L \geq \log_\delta\left( \frac{(1-\delta) \epsilon}{2C} \right)$,
\begin{align}
\mP\left( \|\Phi_{\mathbf{X}}-\hatt{\Phi}_x\|_{max} >\epsilon \right)    \leq p^2 \exp\left( -(N-L) \min\left\{ \frac{\epsilon_1^2}{32(2L+1)^2 n^2C^2 },\frac{\epsilon_1}{8(2L+1) nC } \right\}\right).
\end{align}
\end{lemma}
\begin{remark}
    In the above theorem, we can take $\epsilon_1=9\epsilon/10$, which gives a good looking result. The following Lemma uses this.
\end{remark}

The following lemma bounds the IPSDM by applying the sufficient condition that $\|\Phi_{\mathbf{X}}-\hatt{\Phi}_x\|_{max}<\epsilon/M^4$ $\implies \|\Phi^{-1}_{\mathbf{X}}-\hatt{\Phi}^{-1}_x\|_{max}<\epsilon$ and applies $\epsilon_1=9\epsilon/10$.
\begin{lemma}
\label{lem:IPSD_bound}
Consider a linear dynamical system governed by \eqref{eq:LDIM}. Suppose that the auto-correlation function $R_x(k)$ satisfies exponential decay, $\| R_x(k) \|_2 \leq C \delta^{-|k|}$ and that there exists $M$ such that $\frac{1}{M} \leq \lambda_{min}(\Phi_\X)\leq \lambda_{max}(\Phi_\X)\leq M$. Then for any $0<\epsilon$ and $L \geq \log_\delta\left( \frac{(1-\delta) \epsilon}{2C} \right)$,
{\footnotesize\begin{align}
\mP\left( \|\Phi^{-1}_x-\hatt{\Phi}^{-1}_x\|_{max} >\epsilon \right)    \leq n^2 \exp\left( -(N-L) \min\left\{ \frac{81\epsilon^2}{3200M^{16}(2L+1)^2 n^2C^2 },\frac{9\epsilon}{80M^4(2L+1) nC } \right\}\right).
\end{align}}

\end{lemma}
\label{}
\textbf{Proof:} The following lemma from \cite{matrix_analysis} is useful in deriving this.
\begin{lemma}
\label{lem:matrix_perturbation}
For any invertible matrices $\A$ and $\B$ with $\|\A^{-1}\|_2\|\B-\A\|_2<1$, we have 

{\footnotesize
\begin{align}	\|\A^{-1}-\B^{-1}\|_2 \leq\|\A^{-1}\|_2\|\A\|_2^{-1}\| \B-\A\|_2 \frac{\kappa(\A) }{1-\kappa(\A) \frac{\|\B-\A\|_2}{\|\A\|_2}},
\end{align}}

where $\kappa(\A)$ is the condition number of $\A$.
\end{lemma}
	
Let $\A=\Phi_{x}(z)$ and $\B=\widehat{\Phi}_{x}(z)$. Notice that $\kappa(\Phi_{x})\leq M^2$. Then, by applying Lemma \ref{lem:matrix_perturbation},
\begin{align}
\|\Phi_{x}^{-1}-\widehat{\Phi}_{x}^{-1}\|_2  \label{eq:inverse_ineq}
&\leq \|\Phi_{x}^{-1}\|_2\|\Phi_{x}\|_2^{-1}\| \widehat{\Phi}_{x}-\Phi_{x}\|_2 \frac{M^2 }{1-M^2 \frac{\|\widehat{\Phi}_{x}-\Phi_{x}\|_2}{\|\Phi_{x}\|_2}},\\
\nonumber
&\leq  \frac{M^4 \|\widehat{\Phi}_{x}-\Phi_{x}\|_2 }{1-M \|\widehat{\Phi}_{x}-\Phi_{x}\|_2} <\epsilon,
\end{align}  
if $\|\widehat{\Phi}_{x}-\Phi_{x}\|_2 < \epsilon/M^4$.
Letting $\epsilon_1=\frac{9\epsilon}{10M^4}$ and $\epsilon\xrightarrow{} \epsilon/M^4$ gives the Lemma statement. 
For any matrix $A$ and vector $\x$, $\|A\x\|_2 \leq \|A\|_2$. Also, $\|A\|_{max}=\max_{i,j} e_i^T A e_j \leq \|A\|_2$. Similarly, there exists a $c_1 \in \mbR$ such that $\|A\|_2 \leq c_1\|A\|_{max}$.  Then the above concentration bounds on the maximum value extends to spectral norm also.

Let $\hatt{\A}, \hatt{\B}$ be the estimated values of $\A,\B$. Then, 
\begin{align*}
    \|\A\B-\hatt{\A}\hatt{\B}\|_2&=\|\A\B-\hatt{\A}(\hatt{\B}+\B-\B)\|_2\\
    &\leq \|(\A-\hatt{\A})\B\|_2+\|\hatt{\A}(\hatt{\B}-\B)\|_2\\
    &\leq \|(\A-\hatt{\A})\B\|_2+\|\A(\hatt{\B}-\B)\|_2+\|(\hatt{\A}-\A)(\hatt{\B}-\B)\|_2\\
    &\leq M\|\A-\hatt{\A}\|_2+M\|\hatt{\B}-\B\|_2+\|\hatt{\A}-\A\|\|\hatt{\B}-\B\|_2\\
    &< M\frac{\epsilon}{3M}+M\frac{\epsilon}{3M}+ \frac{\sqrt{\epsilon}\sqrt{\epsilon}}{3}.
\end{align*}
For our Wiener coefficient estimation, $\A$ is $\Phi_\X$ and $\B$ is $\Phi_\X^{-1}$. This gives us a concentration bound on the estimation error of Wiener coefficients using Lemma \ref{lem:PSD_bound} and Lemma \ref{lem:IPSD_bound}.  Notice that $\|\A-\hatt{\A}\|<\epsilon/(3M^5)$ implies that $\|\A\B-\hatt{\A}\hatt{\B}\|_2<\epsilon$.

Therefore
{\footnotesize\begin{align}
\mP\left( \|W_i-\hatt{W}_i\|>\epsilon \right)    \leq n^2 \exp\left( -(N-L) \min\left\{ \frac{81\epsilon^2}{3200c_1^2M^{16}(2L+1)^2 n^2C^2 },\frac{9\epsilon}{80c_1M^4(2L+1) nC } \right\}\right).
\end{align}}

\hfill $\blacksquare$

\section{Primer on Wiener Projection of Time-Series and its Application in Parameter Identification}
\label{app:Wienerproj_timeseries}\
Here, we provide some results on Wiener projection and Wiener coefficients and shows how the Wiener coefficients are useful in system identification. 

Suppose $X_1,\dots,X_n$ be stochastic processes. Consider the  projection of $X_1$ to $X_{\overline{1}}$,
\begin{align}
\label{eq:proj_Wiener_i}
   \Pi_{X_{\overline{1}}}(x_1) :=\arg\min_{\beta\in \mbC^{n-1}} \|X_1-\beta X_{\overline{1}}\|^2=\W_{1\cdot\overline{1}} \X_{\overline{1}}, 
\end{align} where $\W_{1\cdot\overline{1}}(\omega):=\Phi_{X_1\X_{\overline{1}}}(\omega)\Phi_{\X_{\overline{1}}}^{-1}(\omega)$, $\X_{\overline{1}}:=[X_2,\dots,X_n]^T$. Let the solution be 
\begin{align}
X_{1\cdot 2,3,\dots,n}(\omega)=\sum_{j=2}^n X_j(\omega)W_{1j\cdot
23\dots(j-1)(j+1)\dots n}(\omega),
\end{align}
where 
\begin{equation}
\label{eq:Wiener_entry_Phi_expression}
W_{1j\cdot23\dots(j-1)(j+1)\dots n}=\Phi_{x_1x_j}\left(\Phi_{X_{\overline{1}}}^{-1}\right)_{[j,:]}. 
\end{equation}

The following result provides a closed form expression for Wiener coefficients in terms of co-factors of $\Phi_\X$.

\begin{prop}
\label{prop:W_in_terms_ofcofactors}
Let $W_{1\overline{1}}(\omega)=\begin{bmatrix}
    W_{12}(\omega)&\cdots &W_{1n}(\omega)
    \end{bmatrix}$, $\Phi_{1\overline{1}}(\omega)=\begin{bmatrix}
    \Phi_{12}(\omega)& \Phi_{13}(\omega)&\cdots &\Phi_{1n}(\omega)
    \end{bmatrix} $, and 
\begin{align*}
\Phi_{\overline{1}\overline{1}}(\omega)=\begin{bmatrix}
    \Phi_{22}(\omega)& \Phi_{23}(\omega)&\cdots &\Phi_{2n}(\omega)\\
    \Phi_{32}(\omega)& \Phi_{33}(\omega)&\cdots &\Phi_{3n}(\omega)\\
    \vdots& \vdots&\ddots &\vdots\\
    \Phi_{n2}(\omega)& \Phi_{n3}(\omega)&\cdots &\Phi_{nn}(\omega)
    \end{bmatrix}.
\end{align*}
Let the co-factor of $\Phi_{ij}(\omega)$ be \(C_{ij}(\omega)\). Then, for any $k\neq 1$,
\begin{align}
\label{eq:W_1k_cofactor}
    W_{1k\cdot 2\dots(k-1)(k+1)\dots n}(\omega)=-\frac{C_{k1}(\omega)}{C_{11}(\omega)}.
\end{align}
\end{prop}
\textbf{Proof: }See Appendix \ref{app:W_in_terms_ofcofactors}. \hfill

\begin{cor}
The coefficient of $X_k$ in the projection of $X_i$ on $\X_{\overline{i}}:=span\{X_1,\dots,X_{i-1},X_{i-1},\dots,,X_{n} \}$ is
\begin{align}
\label{eq:W_ik_cofactor}
    W_{ik\cdot 123\dots(i-1)(i+1)\dots(k-1)(k+1)\dots n}=-\frac{C_{ki}}{C_{ii}}.
\end{align} 
\end{cor}


\begin{remark}
\label{rem:n=3_Wiener_coeff}
For $n=3$, 
\begin{align}
\label{eq:n=3_Wiener_coeff}
 W_{12\cdot 3}=-\frac{C_{21}}{C_{11}} =\frac{\Phi_{12}\Phi_{33}-\Phi_{31}\Phi_{32}}{\Phi_{22}\Phi_{33}-\Phi_{23}\Phi_{32}}.
\end{align}
\end{remark}

The following lemma shows the linear/conjugate-linear property of the PSD, which is used later in proving single-door criterion.
\begin{lemma}[Linearity of PSD]\
\label{lemma:PSD_linearity}
\begin{enumerate}   
\item Linear on the first term
\begin{align}
\nonumber
    \Phi_{(\alpha X_1+\beta X_2) X_3}=\alpha \Phi_{X_1 X_3}+\beta\Phi_{ X_2 X_3}.
\end{align}
\item  Conjugate-linear on the second term
\begin{align}
\nonumber
    \Phi_{X_3(\alpha X_1+\beta X_2)}=\alpha^* \Phi_{X_3 X_1}+\beta^*\Phi_{ X_3 X_2}.
\end{align}
\end{enumerate}

\end{lemma}

\subsection{Parameter Identification}
\label{subsec:parameter_ID}
To demonstrate the application of Wiener projections in system identification, consider the following LDIM corresponding to Figure \ref{fig:SEM2}: $X_1=\beta(\omega)X_2+\varepsilon_1,$ $X_2=\alpha(\omega) X_3+\varepsilon_2$, and $X_3=\varepsilon_3$, where $\varepsilon_1,\varepsilon_2$, and $\varepsilon_3$ are mutually uncorrelated WSS processes and $X_1,\dots,X_3$ are the observed time series. 
\begin{figure}[!htb]
    \centering
\begin{tikzpicture}[node distance=2.75cm,>=stealth']

\node[vertex style1=blue,xshift=2em] (n3) {$X_3$};

\node[vertex style1=blue, below left=1.5cm of n3,yshift=4ex] (n1) {$X_1$}
edge [<-,cyan!60!blue] node[text style]{$\gamma$} (n3); 

\node[vertex style1=blue, below right=1.5cm of n3,yshift=4ex] (n2) {$X_2$}
    edge [->,cyan!60!blue] node[text style]{$\beta$} (n1)
    edge [<-,cyan!60!blue] node[text style]{$\alpha$} (n3); 

\node[above =0cm of n3,yshift=2ex] (e3) {$\varepsilon_3$}
    edge [->,magenta!60!blue] node[text style]{} (n3); 
\node[left =0cm of n1,xshift=-2ex] (e1) {$\varepsilon_3$}
    edge [->,magenta!60!blue] node[text style]{} (n1); 
\node[right =0cm of n2,xshift=2ex] (e2) {$\varepsilon_2$}
    edge [->,magenta!60!blue] node[text style]{} (n2); 
\end{tikzpicture}

    \caption{Example LDIM with $W_{12\cdot 3}(\omega)=\beta(\omega), \forall(\omega) \in \Omega$.}
    \label{fig:SEM2}
\end{figure}
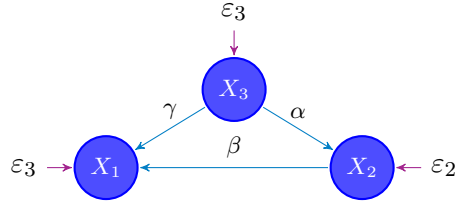

Applying Lemma \eqref{lemma:PSD_linearity}($\omega$ is omitted to avoid cluttering), 
\begin{align*}
    \Phi_{23}&=\Phi_{X_2X_3}=\Phi_{(\alpha X_3+\varepsilon_2)X_3}=\alpha\Phi_{X_3}, \\
    \Phi_{13}&=\Phi_{(\beta X_2+\gamma X_3+\varepsilon_1)X_3}=\Phi_{(\beta (\alpha X_3+\varepsilon_2)+\gamma X_3+\varepsilon_1)X_3}\\
    &\hspace{0cm}=\beta\alpha\Phi_{ X_3}+\gamma \Phi_{ X_3}, \\
    \Phi_{12}&=\Phi_{(\beta X_2+\gamma X_3+\varepsilon_1)X_2}=\beta\Phi_{X_2}+\gamma\Phi_{X_3X_2}\\&=\beta\Phi_{X_2}+\alpha \gamma\Phi_{X_3}.
\end{align*}
Consider the coefficient $W_{12\cdot 3}$ of $X_2$, in the projection of $X_1$ to $span(X_2,X_3)$. From \eqref{eq:n=3_Wiener_coeff}, $ W_{12\cdot 3}(\omega)=\beta(\omega)$, for every $\omega \in \Omega$.
Thus, we can estimate $\beta(\omega)$ from the partial Wiener coefficient $W_{12\cdot3}(\omega)$, for every $\omega\in\Omega$. However, this doesn't work for all the LDIMs. 


\begin{figure}
    \centering
\begin{tikzpicture}[node distance=2.75cm,>=stealth']
\node[vertex style1=blue,xshift=2em] (Z) {$Z$};
\node[vertex style1=blue, right=1.5cm of Z,yshift=0ex] (Y) {$Y$}
edge [<-,cyan!60!blue] node[text style]{$b$} (Z); 
\node[vertex style1=blue, right=1.5cm of Y,yshift=0ex] (X) {$X$}
    edge [<-,cyan!60!blue] node[text style]{$a$} (Y); 
    
\node[above =0cm of Z,yshift=2ex] (e3) {$\varepsilon_Z$}
    edge [->,magenta!60!blue] node[text style]{} (Z); 
\node[above =0cm of Y,yshift=2ex] (e1) {$\varepsilon_Y$}
    edge [->,magenta!60!blue] node[text style]{} (Y); 
\node[above =0cm of X,yshift=2ex] (e2) {$\varepsilon_X$}
    edge [->,magenta!60!blue] node[text style]{} (X); 
\end{tikzpicture}

    \caption{SEM 1}
    \label{fig:SEM1}
\end{figure}

Consider the LDIM in Figure \ref{fig:SEM1}, described by $X=a(\omega)Y+\varepsilon_X$, $Y=b(\omega)Z+\varepsilon_Y$, and $Z=\varepsilon_Z.$. Similar to the above computation, the Wiener coefficient of projecting $Y$ to $X$ and $Z$ can be written as
\begin{align*}
    W_{YZ\cdot X}(\omega)&=\frac{\Phi_{YZ}\Phi_{XX}-\Phi_{YX}\Phi_{ZX}}{\Phi_{ZZ}\Phi_{XX}-\Phi_{XZ}\Phi_{ZX}}\\
    &=b(\omega) \frac{\Phi_Z(\omega)}{|a(\omega)|^2 \Phi_Y(\omega)+\Phi_Z(\omega)}.
\end{align*}
Thus, $W_{YZ\cdot X}(\omega)\neq b(\omega)$ if $a(\omega)\neq 0$. That is, Wiener coefficients do not always give the System parameter directly. However, this is possible if some additional properties are satisfied. For example, projecting $Z$ to $Y$ alone will return the parameter $b(\omega)$ here.

\subsection{Motivating Example for Back-door criterion}
\begin{ex}
\label{ex:causal_effect}
\begin{figure}[htb!]
    \centering
\resizebox{0.6\linewidth}{!}{%
\begin{tikzpicture}[node distance=2.75cm,>=stealth']

\node[vertex style1=blue,xshift=2em] (x1) {$X_{1}$};

\node[vertex style1=blue, right=1.5cm of x1,yshift=0ex,xshift=2em] (x2) {$Y$};

\node[vertex style1=blue, below left=1.5cm of x1,yshift=4ex] (x3) {$X_{3}$}
edge [<-,black!60!blue] node[text style]{} (x1); 

\node[vertex style1=blue, below right=1.5cm of x1,yshift=4ex] (x4) {$X_{4}$}
    edge [<-,black!60!blue] node[text style]{} (x1) 
    edge [<-,black!60!blue] node[text style]{} (x2)
    edge [<-,black!60!blue] node[text style]{} (x3); 

\end{tikzpicture}}
\caption{Graph 1}
\label{fig:example_intervention}
\end{figure}

Consider the graph in Fig. \ref{fig:example_intervention}. The pdf can be factorized as 

{\footnotesize\begin{align*}
f^{(\omega)}\left(x_{1}(\omega),x_{2}(\omega),x_{3}(\omega),x_{4}(\omega) \right)= f^{(\omega)}\left(x_{1} \right)f^{(\omega)}\left(x_{2}\right)
f^{(\omega)}\left(x_{3} \mid x_{1} \right)f^{(\omega)}\left(x_{4} \mid x_{1} x_{2}x_{3} \right).
\end{align*}}
Using \eqref{eq:do_i_definition}, the control distribution on intervention at node $3$ is written as

{\footnotesize \begin{align*}
f^{(\omega)}\left(x_{1},\dots,x_{4} \mid do(x_{3})\right)&= f^{(\omega)}\left(x_{1}(\omega) \right)
    f^{(\omega)}\left(x_{2}(\omega)\right)f^{(\omega)}\left(x_{4}(\omega) \mid x_{1}(\omega) x_{2}(\omega) do(x_{3}(\omega)) \right).
\end{align*}}
Thus, the causal effect of $X_{3}=x_{3}$ on $X_{4}=x_{4}$ is given by (from Theorem \ref{thm:Causal_effect})
{\footnotesize\begin{align*}
    f^{(\omega)}\left(x_{4} \mid do(x_{3})\right)&= \int f^{(\omega)}\left(x_{1}(\omega) \right)f^{(\omega)}\left(x_{2}{(\omega)}\right) \\
    & \hspace{1cm}f^{(\omega)}\left(x_{4}(\omega) \mid x_{1}(\omega) x_{2}(\omega)~do(x_{3}(\omega)) \right) d\nu(x_{1}(\omega) \times x_{2}(\omega)).
\end{align*}}
\end{ex}}

\subsection{Proof of Proposition \ref{prop:W_in_terms_ofcofactors}}
\label{app:W_in_terms_ofcofactors}
Let $\eta_{1\cdot2,3,\dots,n}:=x_1-\sum_{k=2}^n W_{1k}x_k$. From projection theorem \cite{materassi_tac12}, it can be that $\<\eta_{1\cdot2,3,\dots,n},x_j\>=0$, for every $j=2,\dots,n$. Expanding the inner-product, 

{\footnotesize\begin{align*}
    \<x_1-\sum_{k=2}^n W_{1k}x_k,x_j\>&=0 \text{ for } j=2,\dots,n, \\
    \int_{\Omega} \Phi_{x_1x_j}(\omega)-\Phi_{\left(\sum_{k=2}^n, W_{1k}x_k\right)x_j}(\omega) ~\rm \frac{d\omega}{2\pi}&=0,\text{ for } j=2,\dots,n, \\
    \int_{\Omega} \Phi_{x_1x_j}(\omega) -\sum_{k=2}^n W_{1k}(\omega)\Phi_{x_k x_j}(\omega) ~\rm \frac{d\omega}{2\pi}&=0,\text{ for } j=2,\dots,n.
\end{align*}}
Solving this for every $\omega$ \cite{materassi_tac12},
\begin{align*}
    \sum_{k=2}^n W_{1k}(\omega)\Phi_{x_k x_j}(\omega)=\Phi_{x_1x_j}(\omega) \text{ for } j=2,\dots,n.
\end{align*}
Ignoring the index $\omega$, this can be represented as \[\Phi_{1\overline{1}}=W_{1\overline{1}}\Phi_{\overline{1}\overline{1}}.\] Assuming $\Phi_{\overline{1}\overline{1}}$ is invertible, \[W_{1\overline{1}}=\Phi_{1\overline{1}}\Phi^{-1}_{\overline{1}\overline{1}}=\Phi_{1\overline{1}}~\frac{1}{det(\Phi_{\overline{1}\overline{1}})} adj\left(\Phi_{\overline{1}\overline{1}}\right).\]
By definition, cofactor of $\Phi_{ij}$ is $C_{ij}=(-1)^{i+j}det\left(\Phi_{\overline{i}\overline{j}}\right)$. Then, $C_{11}=det\left(\Phi_{\overline{1}\overline{1}}\right)$ and \[W_{1\overline{1}}=\frac{1}{C_{11}}\Phi_{1\overline{1}} adj\left(\Phi_{\overline{1}\overline{1}}\right).\]
The result follows.
\hfill $\blacksquare$

\subsection{Proof of Lemma \ref{lemma:linearity_Wiener_coeff}}
\label{app:linearity_Wiener_coeff}
We prove this for $m=2$. General $m$ follows by induction.

Suppose $y=\alpha_1y_1+\alpha_2y_2$, $\alpha_1,\alpha_2 \in \mF$ and let $\widetilde{\z}:=[x~\z]$. By Wiener projection, $\Pi_{\widetilde{Z}}(y)=\W_{y\widetilde{\z}}\widetilde{\z}$, where $\W_{y\widetilde{\z}}=\Phi_{y\widetilde{\z}}\Phi_{\widetilde{\z}}^{-1}$. Recall that $\Phi_{y\widetilde{\z}}=[\Phi_{yx}~ \Phi_{y{\z}}]$, and from \eqref{eq:Wiener_entry_Phi_expression}, $W_{yx\cdot \z}=b_1^T\Phi_{y\widetilde{\z}}\left(\Phi_{\widetilde{\z}}^{-1}\right)$. Moreover, by linearity (Lemma \ref{lemma:PSD_linearity}), 
$\Phi_{y{\widetilde{\z}}}=\alpha_1\Phi_{y_1\widetilde{\z}}+\alpha_2\Phi_{y_2\widetilde{\z}}$. Thus,
\begin{align*}
\Phi_{y\widetilde{\z}}\Phi_{\widetilde{\z}}^{-1}\widetilde{\z}
&=\alpha_1\Phi_{y_1\widetilde{\z}}\Phi_{\widetilde{\z}}^{-1}\widetilde{\z}~+~ \alpha_2\Phi_{y_2\widetilde{\z}}\Phi_{\widetilde{\z}}^{-1}\widetilde{\z}\\
&=\alpha_1[\Phi_{y_1x}~\Phi_{y_1{\z}}]\Phi_{\widetilde{\z}}^{-1}\widetilde{\z}+\alpha_2[\Phi_{y_2x}~\Phi_{y_2{\z}}]\Phi_{\widetilde{\z}}^{-1}\widetilde{\z}.
\end{align*}
The proof follows since $W_{y_ix\cdot \z}=b_1^T\Phi_{y_i\widetilde{\z}}\left(\Phi_{\widetilde{\z}}^{-1}\right)$ for $i=1,2$. 
\hspace{4cm} \hfill

\section{A Primer on Stochastic Processes and Conditional Independence Notion in Time Domain}
\label{app:SP}
Let $(\Theta,\mM,\mP)$ be a probability space, where $\Theta$ is a sample space, $\mM$ is a sigma field, and $\mP$ is a probability measure. Let $I$ be an index set and let $(S,\Sigma)$ be a measurable space. 

\begin{defn}
A stochastic process (SP) $\X:=\{X(t):t\in I\}$ is a collection of random variables $X(t)$, indexed by $I$, where $X(t):(\Theta,\mM) \mapsto  (S,\Sigma)$. That is, $\{X(t)\}$ takes values in a common space $(S,\Sigma)$. 
The expected value of $X(t)$ is defined as $\mathbb{E}[X(t)]:=\int_{\Theta} X(t) d\mP(\theta) $.
\end{defn}

\begin{remark}
When $S=\mbR^n$ and $\Sigma=\mB_{\mR^n}$, the Borel sigma field, $X(t)$ is a random vector and can be written as $X(t)=[X_{1}(t),\dots,X_{n}(t)]^T$, where $X_{i}(t): (\Theta,\mM) \mapsto (\mbR, \mB_\mbR)$.
\end{remark}

\begin{defn}
Consider a stochastic process $\X:\Theta \mapsto S^I$, defined on the probability space $ \pspace{}.$ The law or the distribution of $\X$, is defined as the push forward measure, $\mu=\mP \circ\X^{-1}$. That is, 
\begin{align*}
    \mu(B):=\mP(\{\theta \in \Theta: \X(\theta )\in B \}), ~\forall B\in \Sigma^I,
\end{align*}
where $\Sigma^I$ denotes the product sigma algebra.
\end{defn}
The distribution of any SP $\X=\{X(t):t\in I\}$ with $I$ non-finite is infinite-dimensional. So, finite-dimensional distributions (FDD) are used to characterize the SPs \cite{CShalizi_lec_notes}.

\begin{defn}[Finite Dimensional Distribution]


Let $\X$ be a stochastic process with distribution $\mu$ as above. The FDD for $t_1,\dots,t_k \in I$ (in arbitrary order), where $k\in \N$, is defined as $\mu_{t_1,\dots,t_k}:=\mP \circ (X(t_1),\dots,X(t_k))^{-1}$.
\end{defn}
\begin{ex}
\label{example:FDD}
Suppose $n=1$ and let $\displaystyle B_k=\bigtimes_{i=1}^k(-\infty,x_{t_i}]$, where $x_{t_1},\dots,x_{t_k} \in \mbR$. Then, $$\mu_{t_1,\dots,t_k}(B_k)=\mP(X(t_1)\leq x_{t_1},\dots,X(t_k)\leq x_{t_k}).$$
\end{ex}
\begin{defn}{(Projection Operator, Coordinate Map):} A projection operator $\pi_t$ is a map from $S^I$ to $S$ such that $\pi_t \X=X(t)$. That is, projection operator provides the marginal distributions, which is obtained by integrating out the rest of the variables from the distribution of $\X$. 

For $i=1,\dots,n$, $\pi_t^{i}: S^I \mapsto S$, defined as $ \pi_t^{(i)}\X =X_{i}(t)$, denote the $i$-th component of $X(t)$. For an index set $J\subset I$, $\mu_J^{(i)}:=\mu \circ \left(\pi_J^{(i)}\right)^{-1}$ is the $i$-th component of the projection of $\mu$ to $J$.

\end{defn}

\begin{defn}[Projective Family of Distributions] A family of distributions $\mu_J$, $J\subset I$, is projective or \textbf{consistent} when for every $J\subset K \subset I$, we have $\mu_J=\mu_K \circ (\pi_J^K)^{-1}$. 
\end{defn}

\begin{lemma} \cite{CShalizi_lec_notes}
The FDDs of a stochastic process always form a projective family. That is, for every $J\subset K \subset I$, we have $\mu_J=\mu_K \circ (\pi_J^K)^{-1}$.
\end{lemma}

\begin{remark} The FDDs form the set $\{\mu_J: J\in Fin( I)\}$, where $Fin(I)$ denotes the collection of all finite subset of $I$. \end{remark}

The existence of the unique probability measure is given by the following theorem.
\begin{lemma} [Daniel-Kolmogorov extension theorem] \cite{CShalizi_lec_notes} \
\label{lem:Daniel-Kolmogorov}
Suppose that we are given, for every $J \subset I$, a consistent family of distributions $\mu_{J}$ on  $(S^J,\Sigma^J)$. Then, there exists a unique probability measure $\mP$ on $(\Theta,\mM)$ such that 
$$\mu_J=\mP \circ \pi_J^{-1}.$$
\end{lemma}
\begin{remark}
The application of Lemma \ref{lem:Daniel-Kolmogorov} is that it is sufficient to study the FDDs, which are easier to handle. Two SPs that have same FDDs have the same distribution \cite{CShalizi_lec_notes}.
\end{remark}

Next, the notion of independence and conditional independence are defined for SPs
\subsection{FDD based Independence in Stochastic Process}

An application of Lemma \ref{lem:Daniel-Kolmogorov} is that the independence can be defined on FDDs.

\begin{defn}[Independence-FDD]
Stochastic processes $\X_1,\dots,\X_m$ are independent if and only if for every $J_1,\dots,J_m \in Fin(I) $,  and $B_{J_i} \in \Sigma^{J_i}$, $i=1,\dots,m$,
\begin{align*}
    \mP\left( \bigcap_{i=1}^m \left\{\pi_{J_i} \X_i\in B_{J_i}\right\} \right)=\prod_{i=1}^m\mP\left(  \pi_{J_i} \X_i\in B_{J_i} \right).
\end{align*}



\end{defn}

\begin{defn}[Conditional independence-FDD]

Consider three stochastic processes $\X_1$, $\X_2$, and $\X_3$ and the finite projection $\pi_{J_i} \X_i$, $i=1,2,3$. $ \X_1$ is said to be conditionally independent of $ \X_2$ given $\X_3$ if and only if for every $J_1,J_2 \in Fin(I) $, there exists $J_3 \in Fin(I)$ such that
\begin{align*}
    &\mP\left( \bigcap_{i=1}^2 \left\{\pi_{J_i} \X_i\in B_{J_i}\right\} \Bigm\vert \pi_{J_3} \X_3\in B_{J_3} \right)=\prod_{i=1}^2\mP\left(  \pi_{J_i} \X_i\in B_{J_i}\Bigm\vert \pi_{J_3} \X_3\in B_{J_3} \right),
\end{align*}
where $B_{J_i} \in \Sigma^{J_i}$, $i=1,\dots,3$.


\end{defn}

\begin{remark}
The similar definition holds for conditional independence of set of stochastic processes also.
\end{remark}

\begin{remark}
    This definition for DAG factorization with FDD will work with the AR processes and the time-series model with independent noise (TiMINo) with finite lags, provided in \cite{peters2013causal}, which include non-linear models. However, FDD based conditional independence notion will fail for infinite convolution model, since a finite $J_3$ might not exist. However, the frequency domain independence will work in the infinite linear convolution model.
\end{remark}

\end{document}